\documentclass{article}

\usepackage{microtype}
\usepackage{graphicx}
\usepackage{subcaption}
\usepackage{booktabs} 

\usepackage{hyperref}

\usepackage[preprint]{styles/icml2026}

\usepackage{amsmath}
\usepackage{amssymb}
\usepackage{mathtools}
\usepackage{amsthm}

\usepackage{url}
\usepackage{graphicx}
\usepackage{color}
\usepackage{xcolor}         
\usepackage{enumitem}
\usepackage{multirow}
\usepackage{colortbl} 
\usepackage{booktabs}
\usepackage{caption}

\usepackage{enumitem}
\usepackage{wrapfig}

\usepackage[capitalize,noabbrev]{cleveref}

\theoremstyle{plain}

\theoremstyle{definition}

\theoremstyle{remark}

\usepackage[textsize=tiny]{todonotes}

\icmltitlerunning{Diagnosing and Mitigating Modality Interference in 
    Multimodal Large Language Models}

\begin{document}

\twocolumn[
    \icmltitle{Diagnosing and Mitigating Modality Interference in \\
    Multimodal Large Language Models}

    \icmlsetsymbol{equal}{*}

  \begin{icmlauthorlist}
    \icmlauthor{Rui Cai}{yyy}
    \icmlauthor{Bangzheng Li}{yyy}
    \icmlauthor{Xiaofei Wen}{yyy}
    \icmlauthor{Muhao Chen}{yyy}
    \icmlauthor{Zhe Zhao}{yyy}
  \end{icmlauthorlist}

  \icmlaffiliation{yyy}{Department of Computer Science, University of California-Davis, Davis, CA, United States}

  \icmlcorrespondingauthor{Rui Cai}{ruicai@ucdavis.edu}
  \icmlcorrespondingauthor{Zhe Zhao}{zao@ucdavis.edu}
  \icmlkeywords{Multimodal Large Language Models, Modality Interference, Robustness}

  \vskip 0.3in
]

\printAffiliationsAndNotice{}

\begin{abstract}
Multimodal Large Language Models demonstrate strong performance on multimodal benchmarks, yet often exhibit poor robustness when exposed to 
spurious modality interference, such as irrelevant text in vision understanding, or irrelevant visual content in question answering.
At its core, modality interference refers to cases where spurious signals from non-essential modalities distort model decisions, which we systematically analyze through causal, perturbation-based diagnostic experiments.
To address this problem, we propose a unified finetuning framework that combines heuristic and adversarial perturbation-based data augmentation with output-level consistency regularization between original and perturbed inputs. Extensive experiments across image-heavy, text-heavy, and multimodal benchmarks, spanning multiple MLLM architectures and model scales, demonstrate consistent improvements in unimodal robustness and generalization, while improving standard multimodal performance.
\end{abstract}

\section{Introduction}

Multimodal Large Language Models (MLLMs) have made significant strides in integrating vision and language understanding within a unified architecture ~\citep{llava, instructblip, qwen25vl}. By combining powerful visual encoders and large language models through alignment mechanisms, MLLMs such as LLaVA~\citep{llava} and Qwen-VL~\citep{qwen25vl} demonstrate strong capabilities across a wide range of multimodal tasks. 
However, beneath their seemingly impressive performance lies a critical limitation: MLLMs often fail to distinguish between task-relevant and task-irrelevant signals across modalities, leading to unreliable predictions~\citep{wang2024mdpo, vlmconflict, spurlens}. A concrete manifestation of this limitation is their behavior in unimodal diagnostic settings, where only a single modality is required for correct reasoning. In such settings, MLLMs frequently underperform on pure visual recognition~\citep{imagewikiqa, mof} and textual reasoning~\citep{vlmconflict, cogvlm, vila}, indicating that non-essential modalities can still influence the prediction process.



Recent studies have attributed 
this failure mode to a variety of learning challenges arising during the multimodal alignment process, such as catastrophic forgetting of pretraining knowledge~\citep{imagewikiqa, mof, cogvlm, vila}, cross-modality knowledge conflict~\citep{wang2024mdpo, vlmconflict}, and 
spurious correlations~\citep{more, spurlens}. 
Catastrophic forgetting has been identified as a key factor in visual degradation~\citep{imagewikiqa, mof}, where multimodal tuning of MLLM overrides its pretrained visual features. Cross-modal knowledge conflict~\citep{wang2024mdpo, vlmconflict} impairs pure-text reasoning, as models often produce inconsistent outputs when visual inputs are introduced, reflecting misaligned visual and textual parametric memories. Additionally, studies on spurious correlations~\citep{more, spurlens, causalmm} show that MLLMs tend to rely on superficial cross-modal cues rather than task-relevant grounding.

Despite these extensive efforts, existing explanations largely remain fragmented, attributing the observed failures to isolated challenges arising during multimodal alignment. As a result, it remains unclear whether these seemingly diverse issues share a common inference-time mechanism that governs how MLLMs identify and rely on task-relevant modalities during prediction. Motivated by this gap, we adopt a unified perspective and argue that the core limitation lies in MLLMs' lack of cross-modality competency~\citep{Competency}—the ability to fairly evaluate and integrate information across modalities. Current models lack mechanisms to support this competency during inference, making them vulnerable to misleading cross-modal signals—a failure mode we refer to as \textbf{\textit{Modality Interference}}.


\begin{figure*}[t]
    \centering
    \includegraphics[width=\linewidth]{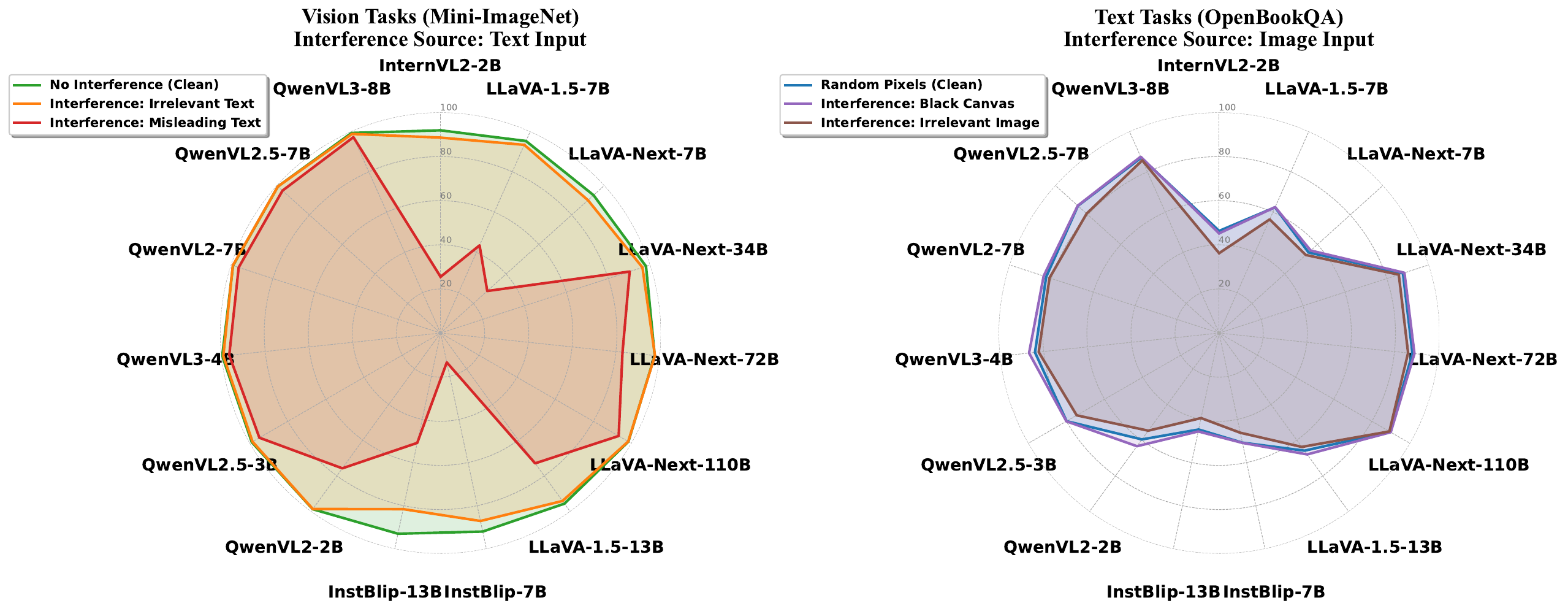}
    \caption{Modality Interference in MLLMs. We visualize the performance of 15 MLLMs using radar charts, where the polygon area signifies model capability.
\textbf{Left (Vision Tasks):} When vision tasks are interfered with by \textit{Misleading Text} (red), models exhibit a severe performance collapse, shrinking towards the center.
\textbf{Right (Text Tasks):} Conversely, text reasoning tasks suffer from \textit{Visual Noise} (brown/irrelevant image), causing a noticeable degradation compared to the no-interference baseline.} 
    \label{fig:pretrained_models_perturbation_results}
    \vspace{-0.3cm}
\end{figure*}


To systematically diagnose and mitigate modality interference, we first design a perturbation-based evaluation experiment inspired by causal intervention principles~\citep{causalIntervention, more}, covering multiple tasks and model scales.
Specifically, in our analysis, we focus on modality-heavy settings using multiple-choice question answering across image-heavy (e.g., image classification), text-heavy (e.g., pure-text QA), and balanced multimodal tasks (e.g., VQA), enabling controlled analysis under different modality-reliance scenarios.
To further induce modality interference, 
we inject heuristic perturbations (e.g., misleading textual descriptions or distractor visual inputs) into task-irrelevant modalities to introduce spurious or misleading signals, and quantify their impact on model predictions.
We evaluate the resulting changes in model predictions to assess the extent of modality interference. While the perturbation-based evaluations offer empirical insights, we further frame our analysis through a causal intervention framework and in which we model this problem through a causal graph abstraction.
Building on this framework, we evaluate a range of pretrained MLLMs and observe consistent failure patterns across tasks and scales, as shown in~\Cref{fig:pretrained_models_perturbation_results}. As illustrated, current MLLMs exhibit significant vulnerability to modality interference: misleading textual descriptions trigger a performance collapse in image-heavy tasks, while irrelevant visual noise disrupts text-heavy reasoning.


The empirical results from~\Cref{fig:pretrained_models_perturbation_results} confirm the presence of modality interference due to their lack of cross-modality competency.
To mitigate this, we propose a perturbation-robust fine-tuning framework. We construct a diverse training mixture incorporating both heuristic noise and adversarial worst-case disruptions to simulate alignment failures, while simultaneously employing consistency regularization (e.g., via Jensen–Shannon divergence) to enforce output stability between clean and perturbed inputs. 
In summary, the main contributions of this paper are threefold. First, we introduce the Cross-Modality Competency Problem to characterize MLLMs’ difficulty in appropriately integrating information across modalities, and identify modality interference as a measurable manifestation of this issue. Second, we propose a causal, perturbation-based evaluation experiment that exposes inappropriate modality reliance and systematically quantifies MLLMs’ susceptibility to task-irrelevant signals. Third, we develop a dataset-agnostic, perturbation-agnostic fine-tuning strategy to mitigate modality interference. Extensive experiments across multiple MLLM families and diverse benchmarks consistently demonstrate the effectiveness and generality of our approach.

\begin{figure*}[h]
    \centering
    \includegraphics[width=\linewidth]{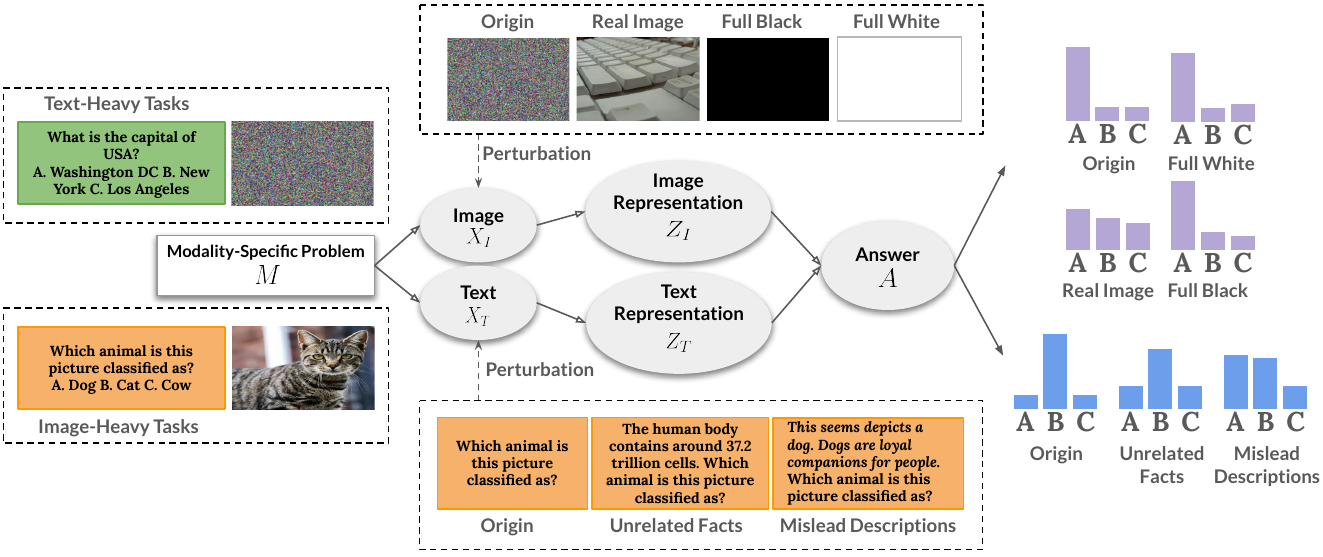}
    \caption{Causal graph illustrating modality interference in our perturbation-based evaluation analysis. Controlled interventions (heuristic) perturb either the image or text inputs, affecting their intermediate representations and ultimately the model prediction.} 
    \label{fig:causal_graph}
    \vspace{-0.3cm}
\end{figure*}

\section{Related Works}

\paragraph{Improving Modality Alignment in Multimodal Language Models}
Recent studies have revealed that modality misalignment remains a key obstacle in MLLMs, leading to degraded performance on both image-heavy and text-heavy tasks. For visual understanding, catastrophic forgetting occurs when multimodal tuning overrides pretrained visual features~\citep{imagewikiqa, mof, cogvlm}. In text-heavy scenarios, knowledge conflict~\citep{vlmconflict} arises when inconsistent parametric knowledge from different modalities confuse reasoning. mDPO~\citep{wang2024mdpo} identifies language bias in training, where models fail to condition their responses on visual input. Some works attribute such issues to shallow fusion~\citep{cogvlm}—e.g., LLaVA~\citep{llava} uses lightweight projectors to bridge vision and language spaces, leaving a representational gap and resulting in loosely coupled features~\citep{mof, vlmconflict, EECA}. Others highlight data limitations: even well-encoded visual features fail to support reasoning without adequate supervision to guide decoding~\citep{imagewikiqa}. Building on these diagnoses, recent works propose multiple solutions. MoF~\citep{mof} mitigates this by fusing features from multiple vision encoders, while VLMClassifier~\citep{imagewikiqa} enhances recognition via vision-only finetuning, though it struggles with VQA due to lack of cross-modal alignment. CogVLM~\citep{cogvlm} introduces a visual expert module to improve vision-language integration. VILA~\citep{vila}, QwenVL~\citep{qwen25vl}, and InternVL~\citep{internvl3} incorporate text-only supervision in different ways to preserve or enhance language capabilities during multimodal training—through stage-wise separation, parallel preservation, and unified joint optimization, respectively. Similar patterns also arise in multimodal structural reasoning~\citep{zrlmmkg}. However, these approaches largely treat modality contributions implicitly, without explicitly modeling how multimodal inputs influence the model’s decision process.

\vspace{-0.5cm}\paragraph{Adversarial Robustness Across Modalities}
Adversarial perturbations threaten the reliability of both vision and text models by exploiting vulnerabilities in continuous image embeddings and discrete token spaces.
In vision tasks, attacks like FGSM~\citep{DBLP:journals/corr/GoodfellowSS14} and CW~\citep{DBLP:conf/sp/Carlini017} first revealed the fragility of neural networks to imperceptible input changes. PGD~\citep{DBLP:conf/iclr/MadryMSTV18} formalized this under a saddle-point framework, becoming the standard for adversarial training. AutoAttack~\citep{DBLP:conf/icml/Croce020a} further unified strong attacks, including PGD variants, into a reliable benchmark. In text tasks, adversarial methods must contend with discrete inputs. TextFooler~\citep{DBLP:conf/aaai/JinJZS20} substitutes key words with semantically similar ones to mislead predictions, while CodeAttack~\citep{DBLP:conf/aaai/JhaR23} adapts this idea to code-language models. More recently, PGD has been extended to LLMs via continuous relaxation~\citep{DBLP:journals/corr/abs-2402-09154}, enabling efficient attacks in embedding space. PTP~\citep{DBLP:conf/emnlp/ChenCHC23} applies PGD-style perturbations in the prompt embedding space to smooth training and enhance stability. Inspired by this, our work extends to the multimodal embedding space, enabling unified gradient-based control over both visual and textual inputs with unique masking strategy.


\section{Causal Analysis on Modality Interference}
\label{sec:causal_experiment_details}

\paragraph{Cross-Modality Competency Problems in Multimodal Large Language Models}

Competency problems describe scenarios where models rely on spurious correlations between isolated input features and output labels to make predictions, instead of leveraging meaningful interactions among multiple features~\citep{Competency}.  
We extend this concept to the multimodal setting by treating entire modalities (e.g., image $X_I$ or text $X_T$) as structured feature sources. 
We define the \textit{Cross-Modality Competency} as an ability for MLLM to 
fairly evaluate and integrate all modalities, identifying which ones carry task-relevant signals while ignoring misleading or irrelevant ones.
In pure-text question answering with MLLMs, the model is provided with both a question and an image, although the image is not required to solve the task. When predictions are influenced by spurious visual cues correlated with certain answers, the model allows irrelevant modality signals to influence its reasoning and predictions, resulting in modality interference. 


\vspace{-0.6cm}\paragraph{Perturbation-based Evaluation Experiment}
To systematically measure cross-modality competency, we propose a perturbation-based evaluation framework. The core idea is to inject controlled noise into the irrelevant modality and assess the model’s robustness to such perturbations. Specifically, for image-heavy tasks, we perturb the text input by: (1)Prepending \textit{unrelated scientific facts}; (2) Prepending \textit{misleading descriptions} that falsely link incorrect options to the image content. For text-heavy tasks, we perturb the visual input by: (1) Attaching a \textit{semantically meaningful real but irrelevant image}; (2) Substituting with a \textit{full black} or \textit{full white} canvas image. 
Models with strong modality selectivity should maintain high prediction consistency when irrelevant modality signals are perturbed. We select multiple image-heavy and text-heavy tasks for evaluation. Each task is framed as a multiple-choice classification problem, requiring the model to choose the correct option based on image and text modalities as input, with perturbations applied as described above. To validate robustness and generalization for our proposed mitigation method, we additionally consider real-world out-of-distribution perturbations reflecting practical noise encountered in deployed multimodal systems: (i) noisy OCR snippets for image-heavy tasks; and (ii) unrelated UI screenshots for text-heavy tasks. Details in \Cref{sec:experiment}, \Cref{apx:Perturbation-based Evaluation Experiment Results},  \Cref{tab:modality_interference_vinilla_models} and \Cref{tab:ood-results-new-perturbations} and \Cref{tab:ood-results-new-datasets}.  

\vspace{-0.4cm}\paragraph{Causal Modeling of Modality Interference}

We formalize modality interference through a causal intervention perspective with a causal graph, as shown in~\Cref{fig:causal_graph}, where visual inputs ($X_I$) and textual inputs ($X_T$) are processed into their respective representations ($Z_I$, $Z_T$) before being fused to produce the final prediction ($A$). To study the model’s reliance on different modalities, we introduce perturbations directly at the input level, serving as causal interventions~\citep{causalIntervention} on $X_I$ and $X_T$. Under ideal cross-modality competency, the model’s prediction should primarily depend on the task-relevant pathway (e.g., $X_I \rightarrow Z_I \rightarrow A$ in image-heavy tasks, $X_T \rightarrow Z_T \rightarrow A$ in text-heavy tasks). Causal interventions at the input level allow us to diagnose whether the model improperly fuses irrelevant signals into its decision process. We use $x_I’$ to denote an intervention on image $X_I$ and use $x_T’$ as an intervention on text $X_T$. Following Pearl’s causal framework~\citep{causalIntervention, intermediaryVariable}, we quantify the impact of modality perturbations on model predictions by formalizing causal effects in our multimodal setting. Specifically, we define the pre-intervention prediction distribution as $P(A|X_I, X_T)$, and the post-intervention prediction distribution after applying a perturbation on $X_I$ or $X_T$ as $P’(A|\text{do}(X_I=x_I’)$ or $\text{do}(X_T=x_T’))$. The \textit{do}-operation represents an intervention to specific modality, and the causal effect (CE) of an intervention is evaluated via a distance metric $\delta$ comparing $P$ and $P’$ as $\text{CE} = \delta(P, P')$.
We assess the causal effect via prediction changes using $\delta_{\text{cp}}(P, P') := \mathbb{I}\left( a \neq a' \right)$
in which $a = \arg\max_x P(x)$ is the predicted answer before intervention, $a’ = \arg\max_x P’(x)$ is the predicted answer after intervention and $\mathbb{I}(\cdot)$ is the indicator function that outputs $1$ if $a \neq a’$ and $0$ otherwise. Thus, $\delta_{\text{cp}}$ captures whether the model’s final decision $A$ changes under perturbations to the input modality. 
In all interventions, a high value of $\delta_{\text{cp}}$ indicates the model’s susceptibility to modality interference, revealing spurious reliance on irrelevant modality.

\section{Methods}

\begin{figure}[h]
\centering
\includegraphics[width=\linewidth]{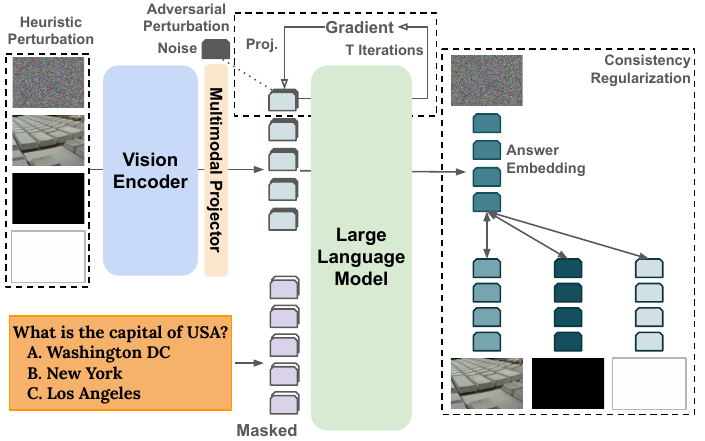}
\caption{Overview of our proposed framework.}
\label{fig:framework_pipeline}
\vspace{-0.5cm}
\end{figure}

To mitigate modality interference and enhance cross-modality competency, we propose a unified perturbation-aware training framework that introduces interventions at both the input level (on $X_I$ and $X_T$) and the representation level (on $Z_I$ and $Z_T$) with consistency regularization. Overall pipeline is displayed in ~\Cref{fig:framework_pipeline}. 

\subsection{Perturbation-based Data Augmentation}

To increase causal robustness along the desired paths $X_I \rightarrow Z_I \rightarrow A$ (for image-heavy tasks) or $X_T \rightarrow Z_T \rightarrow A$ (for text-heavy tasks), we first adopt a causally grounded data augmentation method by augmenting each sample with both heuristic perturbations and training time adversarial perturbations.

\vspace{-0.3cm} \paragraph{Mixture of Multi-Task Training data with Heuristic Perturbations}
Let $\mathcal{D} = {(x_I, x_T, a)}$ denote the full multimodal training dataset, where $(x_I, x_T)$ are the image and text inputs and $a$ is the ground-truth answer. The dataset $\mathcal{D}$ can be partitioned into three subsets based on tasks: 1) Image-heavy set $\mathcal{D}^{\text{img}}$: samples where visual input $x_I$ is the dominant information source; 2) Text-heavy set $\mathcal{D}^{\text{text}}$: samples where textual input $x_T$ provides the main reasoning signal; 3) VQA set $\mathcal{D}^{\text{vqa}}$: samples from vision-language datasets with naturally balanced multimodal dependencies. In practice, we transform these image-heavy and text-heavy datasets into VQA format to construct $\mathcal{D}^{\text{img}}$ and $\mathcal{D}^{\text{text}}$, and derive VQA samples from the supervised finetuning stage of each MLLM as $\mathcal{D}^{\text{vqa}}$. For each sample $(x_I, x_T, a) \in \mathcal{D}^{\text{img}} \cup \mathcal{D}^{\text{text}}$, we maintain its original version and apply heuristic perturbation to construct Origin Samples and Perturbation-Augmented Samples. Origin samples are used to reinforce the desired causal path (e.g., $X_I \rightarrow Z_I \rightarrow A$). Perturbation-augmented samples are variants of the same instance with perturbations applied to the irrelevant modality, which are denoted as $(x_I, \tilde{x}_T, a) \in \mathcal{D}^{\text{img}}_{\text{pert}}$ and $(\tilde{x}_I, x_T, a) \in \mathcal{D}^{\text{text}}_{\text{pert}}$ where $\tilde{x}_T$ and $\tilde{x}_I$ are perturbed text and image respectively. Together, the augmented dataset can be written as: 
\begin{equation}
\label{eqa:data_consititution}
    \mathcal{D}^{\text{AUG}} = \mathcal{D}^{\text{img}} \cup \mathcal{D}^{\text{img}}_{\text{pert}} \cup \mathcal{D}^{\text{text}} \cup \mathcal{D}^{\text{text}}_{\text{pert}} \cup \mathcal{D}^{\text{VQA}}.
\end{equation}
We sample $N_{\text{img}}$ and $N_{\text{text}}$ examples from $\mathcal{D}^{\text{img}}$ and $\mathcal{D}^{\text{text}}$ to construct $\mathcal{B}^{\text{img}}_{\text{orig}}$ and $\mathcal{B}^{\text{text}}_{\text{orig}}$ respectively, and the remaining $N_{\text{vqa}}$ examples are VQA samples from $\mathcal{D}^{\text{vqa}}$ to construct $\mathcal{B}^{\text{vqa}}$ . With dynamically generated perturbed variants for each sample, the final training batch is:
\begin{equation}
    \mathcal{B} = \mathcal{B}^{\text{img}}_{\text{orig}}
\cup \mathcal{B}^{\text{img}}_{\text{pert}}
\cup \mathcal{B}^{\text{text}}_{\text{orig}}
\cup \mathcal{B}^{\text{text}}_{\text{pert}}
\cup \mathcal{B}^{\text{vqa}},
\end{equation}
where $\mathcal{B}^{\text{img}}_{\text{pert}}$ and $\mathcal{B}^{\text{text}}_{\text{pert}}$ are perturbation-augmented variants generated from $\mathcal{B}^{\text{img}}_{\text{orig}}$ and $\mathcal{B}^{\text{text}}_{\text{orig}}$, respectively. 
For the full training batch $\mathcal{B}$, which includes both original and perturbed samples, we define the supervised loss $\mathcal{L}_{\text{sft}}$ as the cross-entropy loss computed over all answer tokens in the ground-truth sequences. Let $\mathcal{L}_{\text{cls}}(x_I, x_T, a)$ denote the standard autoregressive loss for a sample $(x_I, x_T, a)$, then:
\begin{equation}
    \mathcal{L}_{\text{sft}} = \frac{1}{|\mathcal{B}|} \sum_{(x_I, x_T, a) \in \mathcal{B}} \mathcal{L}_{\text{cls}}(x_I, x_T, a).
\end{equation}

\vspace{-0.4cm}\paragraph{Adversarial Perturbation with Cross-Modality Masking} 
While heuristic perturbations simulate realistic but limited modality noise at the input level, they may not fully capture the worst-case failure modes of MLLMs, especially under complex spurious alignments in the representation space. To overcome this limitation, we introduce a stronger and more generalizable intervention through adversarial training. 
These perturbations simulate worst-case alignment disruptions during training, serving as targeted interventions on latent nodes $(Z_I, Z_T)$ to reveal the \textit{Direct Causal Effect} (DCE) of irrelevant modalities on $A$. By optimizing the model under such adversarial conditions, we reduce the model's reliance on spurious cross-modal signals and reinforce task-relevant causal pathways.
Inspired by PGD~\citep{DBLP:conf/iclr/MadryMSTV18, ptp}, we design a tailored perturbation strategy for multimodal token embeddings $(Z_I, Z_T)$. 
Unlike standard PGD that applies coarse sign-based updates, our method introduces two critical modifications:  
(1) \textit{Modality-specific perturbation masking}, which restricts perturbations to task-irrelevant modalities via a binary mask, thereby transforming noise into targeted causal probes rather than indiscriminate corruption.  
(2) \textit{Raw-gradient updates}, where we remove the sign operator and apply the raw gradient directly, yielding smoother, more diverse, and more realistic perturbations that better simulate modality interference.  
Formally, we construct perturbations $\delta = (\delta_I, \delta_T)$ in the latent space that maximize the model’s predictive loss:
\begin{equation}
    \delta = \underset{ \|\delta\|_\infty \leq \epsilon }{\text{argmax}}\, \mathcal{L_\text{cls}}(f(Z_I + \delta_I, Z_T + \delta_T)),
\end{equation}
where $\epsilon$ bounds the perturbation strength and $f$ is the prediction function. We optimize $\delta$ through $n$ raw-gradient steps, updating at each step $t$ as:
\begin{equation}
    \delta^{(t+1)} = \Pi_{\|\delta\|_\infty \leq \epsilon} \Big( \delta^{(t)} + \alpha \cdot \nabla_{\delta} \mathcal{L_\text{cls}}(f(Z + \delta^{(t)})) \Big),
\end{equation}
where $\alpha$ is the step size, $\Pi$ projects the noise into the $\ell_\infty$ ball, and $Z = [Z_I; Z_T]$ is the concatenated embedding. 
In practice, we integrate a modality-specific binary mask $M \in \{0,1\}^{L \times d}$, where $L$ is the sequence length and $d$ the hidden dimension, ensuring that perturbations only affect task-irrelevant tokens. Given multimodal embeddings $\mathbf{E} \in \mathbb{R}^{L \times d}$, the perturbed embeddings are:
\begin{equation}
    \tilde{\mathbf{E}} = \mathbf{E} + \delta \odot M,
\end{equation}
with $\odot$ denoting element-wise masking. We initialize $\delta$ with Gaussian noise $\mathcal{N}(0,\epsilon^2)$ and update it for $T$ steps. The final adversarial objective is $\mathcal{L}_{\text{adv}} = \mathcal{L}_\text{cls}(f(\tilde{\mathbf{E}}))$.

\subsection{Consistency Regularization under Perturbations}

While perturbation-based augmentation exposes the model to diverse interventions, it does not constrain how intermediate representations ($Z_I$, $Z_T$) should respond, and even small changes in the task-irrelevant modality may cause undesirable shifts in fused features. To address this, we introduce a consistency regularization strategy that enforces \textit{output stability} between original and perturbed inputs, serving as an indirect constraint on $Z_I$ and $Z_T$ to mitigate modality interference. By minimizing the divergence between the prediction distributions of original and perturbed inputs, the model is encouraged to maintain invariant behavior along the task-relevant causal paths.  
Formally, given an original input $x$ and its perturbed counterpart $\tilde{x}$, with predictive distributions $p_\theta(A|x)$ and $p_\theta(A|\tilde{x})$, the consistency loss follows the general form:
\begin{equation}
\mathcal{L}_{\text{consistency}} = \mathrm{Consistency}\big( p_\theta(A|x) \,\|\, p_\theta(A|\tilde{x}) \big).
\end{equation}
In practice, we instantiate this by applying distributional divergence (e.g., KL or JS) at the token level. Let $l^{\text{orig}}, l^{\text{pert}} \in \mathbb{R}^{L_A \times V}$ denote the pre-softmax logits for the original and perturbed samples, where $L_A$ is the number of answer tokens and $V$ the vocabulary size. Using KL divergence with temperature $\tau$ as an example, the loss becomes (equally apply to image-heavy and text-heavy tasks):
\begin{equation}
\begin{small}
\mathcal{L}_{\text{consistency}} = \frac{1}{L_A} \sum_{i=1}^{L_A} \sum_{v=1}^{V} 
\text{softmax}\!\left(\tfrac{l^{\text{orig}}_i}{\tau}\right)_v
\cdot 
\log \frac{\text{softmax}\!\left(\tfrac{l^{\text{orig}}_i}{\tau}\right)_v}
{\text{softmax}\!\left(\tfrac{l^{\text{pert}}_i}{\tau}\right)_v}.
\end{small}
\end{equation}

\begin{table*}[t]
\centering
\caption{Evaluation on unimodal and VQA datasets. For unimodal datasets, we report accuracy under the original input (Orig) and the worst-performing perturbation (Perturbed). For VQA datasets, we report accuracy on original setting. Best results are highlighted in bold.}
\vspace{-0.6em}
\label{tab:unimodal_vqa_combined_baselines}
\resizebox{\textwidth}{!}{
\begin{tabular}{l|cc|cc|cc|cc|c|c|c}
\toprule
\multirow{2}{*}{\textbf{Model Settings}} &
\multicolumn{2}{c|}{\textbf{Mini-ImageNet}} &
\multicolumn{2}{c|}{\textbf{Caltech-101}} &
\multicolumn{2}{c|}{\textbf{OpenBookQA}} &
\multicolumn{2}{c|}{\textbf{MMLU}} &
\multicolumn{1}{c|}{\textbf{ScienceQA}} &
\multicolumn{1}{c|}{\textbf{MM-Bench}} &
\multicolumn{1}{c}{\textbf{Seed-Bench}} \\
& Orig & Perturbed & Orig & Perturbed & Orig & Perturbed & Orig & Perturbed & Accuracy & Accuracy & Accuracy \\
\midrule
LLaVA-1.5-7B       & 95.3 & 43.5 & 97.0 & 57.4 & 62.4 & 56.4 & 46.3 & 45.2 & 64.5 & 64.3 & 63.4 \\
\quad+ CoT~\citep{cot}          & 81.7 & 28.9 & 80.9 & 36.0 & 38.8 & 38.9 & 39.6 & 38.7 & 64.7 & 65.2 & 64.1 \\
\quad+ VLMClassifier-1~\citep{imagewikiqa}     & 15.1 & 0.0 & 15.6 & 0.0 & 61.5 & 61.5 & 47.8 & 47.5 & 61.1 & 36.2 & 35.9 \\
\quad+ VLMClassifier-2~\citep{imagewikiqa}   & 15.6 & 0.0 & 15.1 & 0.0 & 61.8 & 61.2 & 47.4 & 47.8 & 61.8 & 35.8 & 36.0 \\
Ours & \textbf{98.6} & \textbf{98.4} & \textbf{99.3} & \textbf{98.9} & \textbf{81.8} & \textbf{81.0} & \textbf{51.5} & \textbf{51.0} & \textbf{67.8} & \textbf{73.1} & \textbf{64.6} \\
\midrule
LLaVA-1.5-13B    & 95.6 & 73.0 & 97.9 & 77.4 & 65.9 & 63.8 & 51.8 & 50.8 & 66.1 & 72.1 & 64.5 \\
\quad+ CoT~\citep{cot}         & 92.9 & 62.8 & 96.6 & 67.8 & 55.6 & 53.0 & 47.3 & 45.5 & 65.6 & 70.2 & 64.9 \\
\quad+ I-MoF~\citep{mof} & 93.9 & 70.1 & 97.8 & 80.9 & 69.2 & 64.5 & 46.2 & 39.1 & \textbf{66.8} & 73.0 & 66.6 \\
Ours & \textbf{98.5} & \textbf{98.4} & \textbf{99.2} & \textbf{98.6} & \textbf{83.0} & \textbf{82.1} & \textbf{56.6} & \textbf{55.8} & 62.6 & \textbf{73.7} & \textbf{68.4} \\
\midrule
Qwen2.5-VL-7B   &  99.3 & 96.3 & 99.1 & 97.2 & 85.9 & 77.5 & 69.3 & 63.7  & 85.1 & 86.6 & 77.0 \\
\quad+ CoT~\citep{cot}  & 99.1 & 91.4 & 98.5 & 92.5 & 77.8 & 68.8 & 61.5 & 57.4 & 85.2 & 86.7 & 77.1 \\
R1-OneVision-7B~\citep{r1onevision} & 98.2 & 87.7 & 96.5 & 84.7 & 81.2 & 77.6 & 63.2 & 60.8 &  84.5 & 85.4 & 76.3 \\
Ours & \textbf{99.6} & \textbf{99.5} & \textbf{99.6} & \textbf{99.7} & \textbf{90.9} & \textbf{89.7} & \textbf{69.8} & \textbf{68.2} & 83.9 & \textbf{87.0} & \textbf{78.3}  \\
\bottomrule
\end{tabular}
}
\vspace{-0.5cm}
\end{table*}

\vspace{-0.5cm}\subsection{Final Training Objective}

Our final training objective integrates both perturbation-based data augmentation and consistency regularization into a unified framework. For each batch, we begin with a set of original samples $\mathcal{B}_{\text{orig}}$ and dynamically construct their heuristic perturbed counterparts $\mathcal{B}_{\text{pert}}$ via input-level augmentations. We then apply adversarial perturbations on both $\mathcal{B}_{\text{orig}}$ and $\mathcal{B}_{\text{pert}}$, and enforce consistency between the original and all perturbed samples. The overall loss is:
\begin{equation}
\mathcal{L}_{\text{total}} = \mathcal{L}_{\text{sft}} + \mathcal{L}_{\text{adv}} + \lambda_{\text{cons}} \cdot \mathcal{L}_{\text{consistency}},
\end{equation}
where $\mathcal{L}_{\text{sft}}$ is the supervised loss computed over $\mathcal{B}$, $\mathcal{L}_{\text{adv}}$ is the adversarial loss computed on all adversarial perturbed samples and $\mathcal{L}_{\text{consistency}}$ is the consistency loss between original and all perturbed sample pairs. 
By aligning all three losses with the causal structure of multimodal reasoning, we systematically mitigate modality interference and improve cross-modality competency in MLLMs.

\begin{table*}[t] 
\centering
\small 
\caption{Multimodal reasoning accuracy (\%) on VQA benchmarks under various ablation study configurations. The best accuracy is marked in bold. Arrows (↑ / ↓) indicate relative changes compared to the Vanilla baseline of each model. Overall performance is computed as a weighted average across datasets, with weights proportional to each dataset’s test size~\Cref{tab:dataset_sizes}. (full results in \Cref{tab:vqa_summary_extra_models})}
\label{tab:vqa_summary}
\begin{tabular}{llcccc}
\toprule
\textbf{Model} & \textbf{Method} & \textbf{ScienceQA-IMG} & \textbf{MM-Bench-EN} & \textbf{Seed-Bench-IMG} & \textbf{VQA Overall} \\
\midrule
\rowcolor{gray!40}
\multicolumn{6}{c}{\textbf{\textit{3B Multimodal Models}}} \\
\midrule
\multirow{6}{*}{\shortstack{\textbf{Qwen2.5-VL-3B}\\\scriptsize\cite{qwen25vl}}}
& Vanilla & 63.0 (100\%) & 81.8 (100\%) &	72.3 (100\%) & 72.5 (100\%) \\
& FFT with $D^{\text{VQA}}$ &  72.2 (↑14.6\%) & 82.0 (↑0.2\%) & 74.7 (↑3.3\%) & 75.5 (↑3.9\%) \\
& FFT with $D^{\text{AUG}}$ & 73.6 (↑16.8\%) & 81.7 (↓0.1\%) & 75.3 (↑4.1\%) & \cellcolor{gray!20}\underline{76.2 (↑5.1\%)}\\
& \quad + KL & 72.8 (↑15.6\%) & 80.9 (↓1.1\%) & 74.9 (↑3.6\%) & 75.9 (↑4.7\%) \\
& \quad + ADV & 73.1 (↑16.0\%) & 81.7 (↓0.1\%) & 75.3 (↑4.1\%) & 76.1 (↑5.0\%) \\
& Ours & 73.8 (↑17.1\%) & 82.1 (↑0.1\%) & 75.5 (↑4.4\%) & \cellcolor{gray!20}\textbf{76.4 (↑5.4\%)}\\
\midrule
\rowcolor{gray!40}
\multicolumn{6}{c}{\textbf{\textit{7B Multimodal Models}}} \\
\midrule
\multirow{6}{*}{\shortstack{\textbf{LLaVA-1.5-7B}\\\scriptsize\cite{llava15}}}
& Vanilla & 64.5 (100\%) & 64.3 (100\%) & 63.4 (100\%) & 63.8 (100\%) \\
& FFT with $D^{\text{VQA}}$ & 61.6 (↓4.5\%) & 71.4 (↑11.0\%) & 62.4 (↓1.6\%) & 64.0 (↑0.3\%) \\
& FFT with $D^{\text{AUG}}$ & 65.4 (↑1.4\%) & 71.1 (↑10.6\%) & 63.9 (↑0.8\%) & 65.6 (↑2.8\%) \\
& \quad + JS & 63.8 (↓1.1\%) & 73.5 (↑14.3\%) & 63.7 (↑0.5\%) & 65.6 (↑2.8\%) \\
& \quad + ADV & 66.7 (↑3.4\%) & 71.4 (↑11.0\%) & 63.6 (↑0.3\%) & \cellcolor{gray!20}\underline{65.7 (↑3.0\%)} \\
& Ours & 67.8 (↑5.1\%) & 73.1 (↑13.7\%) & 64.6 (↑1.9\%) & \cellcolor{gray!20}\textbf{66.8 (↑4.7\%)} \\
\midrule
\rowcolor{gray!40}
\multicolumn{6}{c}{\textbf{\textit{13B Multimodal Models}}} \\
\midrule
\multirow{6}{*}{\shortstack{\textbf{InstructBLIP-13B}\\\scriptsize\cite{instructblip}}}
& Vanilla                   & 65.8 (100\%) & 72.1 (100\%) & 63.8 (100\%) & 65.8 (100\%) \\
& FFT with $D^{\text{VQA}}$ & 61.7 (↓6.2\%) & 68.5 (↓5.0\%) & 56.0 (↓12.2\%) & 59.4 (↓9.7\%) \\
& FFT with $D^{\text{AUG}}$ & 66.6 (↑1.2\%) & 71.5 (↓0.8\%) & 64.0 (↑0.3\%) & 65.9 (↑0.2\%) \\
& \quad + KL                & 67.2 (↑2.2\%) & 72.6 (↑0.7\%) & 64.0 (↑0.3\%) & 66.2 (↑0.6\%) \\
& \quad + ADV               & 66.1 (↑0.5\%) & 74.0 (↑2.6\%) & 64.1 (↑0.5\%) & \cellcolor{gray!20}\underline{66.4 (↑0.9\%)} \\
& Ours                      & 66.2 (↑0.6\%) & 73.2 (↑1.5\%) & 64.3 (↑0.7\%) & \cellcolor{gray!20}\textbf{66.5 (↑1.1\%)} \\
\bottomrule
\end{tabular}
\vspace{-0.4cm}
\end{table*}

\section{Experiments}

\label{sec:experiment}

\textbf{Models.} We conduct experiments on three MLLM families with different parameter size: Qwen2.5-vl-3b~\citep{qwen25vl}, LLaVA-1.5-7B \& LLaVA-1.5-13B~\citep{llava15} and InstructBLIP-Vicuna-7B~\citep{instructblip}. 
Following LLaVA and Qwen-VL, we freeze the vision encoder and train the multimodal projector and language model; for InstructBLIP, we instead freeze both the vision encoder and Q-Former, fine-tuning only the projection layer and language model (see \Cref{Apx:Finetuning Strategy}).


\vspace{-0.2cm}\textbf{Baselines.} We include following baselines for comparison: \textit{LLaVA-1.5-13B + I-MoF}~\citep{mof}: By applying the designed Interleaved Mixture-of-Features (I-MoF) module on LLaVA-1.5-13B to spatially combine CLIP~\citep{clip} and DINOv2~\citep{dinov2} visual tokens, it enhances visual grounding by integrating complementary features from contrastive and self-supervised vision encoders. \textit{VLMClassifier}~\citep{imagewikiqa}: it enhances visually-grounded language models for image classification by fine-tuning them on ImageNet~\citep{imagenet} (\textit{VLMClassifier-1}) or ImageNet combining LLaVA-Instruct~\citep{llava} (\textit{VLMClassifier-2}). 
\textit{R1-OneVision-7B}~\citep{r1onevision}: a reasoning-enhanced ``thinking'' model built upon the Qwen2.5-VL-7B backbone. We include it to investigate whether intrinsic reasoning capabilities can inherently mitigate modality interference.
\textit{Chain-of-Thought (CoT) Prompting}~\citep{cot}: we further evaluate prompt-based mitigation by encouraging structured reasoning through CoT-style prompting. The specific prompt design and results are reported in \Cref{sec:cot}. 

\vspace{-0.2cm}\textbf{Datasets.} We evaluate models on benchmarks covering three task types:  \textit{(i) Image-heavy tasks}: Mini-ImageNet~\citep{miniimagenet} and Caltech-101~\citep{caltech101}, used in both training and evaluation; \textit{(ii) Text-heavy tasks}: OpenBookQA~\citep{openbookqa} and MMLU~\citep{mmlu}, used in both training and evaluation; \textit{(iii) VQA tasks}: For training, we use LLaVA-Instruct-dataset~\citep{llava} as the instruction-tuning dataset for related models. For InstructBLIP, we additionally use TextCaps~\citep{textcaps} following its original training configuration. For Qwen2.5-VL, whose instruction-tuning data is proprietary, we adopt LLaVA-Instruct as a standardized alternative. For evaluation, we adopt three multiple-choice VQA benchmarks: ScienceQA-IMG~\citep{scienceqa}, MM-Bench-EN~\citep{mmbench}, and Seed-Bench-IMG~\citep{seedbench}. For ScienceQA and Seed-Bench, we only include examples with image context. For MM-Bench, we use the English version. All datasets are converted into a unified multiple-choice VQA format. We report the accuracy of model predictions and quantify the causal effect with the prediction change rate $\delta_{\text{cp}}$. All models are fine-tuned for 1 epoch with a fixed batch size $N_{\text{batch}}$. All the hyperparameters 
are listed in \Cref{Apx:Parameter Setting}. For each dataset, results are averaged over multiple independent runs. Following standard practice, inference is performed with deterministic decoding (temperature fixed at 0).


\vspace{-0.4cm}\paragraph{Achieving Pareto-Optimality Across Unimodal and Multimodal Tasks}
As shown in~\Cref{tab:unimodal_vqa_combined_baselines}, our method outperforms all baselines across different base MLLMs, demonstrating stronger robustness to modality interference. 
While CoT slightly improves performance in certain VQA settings, its overall gains are minimal and inconsistent, and it fails to mitigate modality interference (e.g., 85.9\% vs.\ 97.9\% on Mini-ImageNet). 
Similarly, even the reasoning-enhanced R1-OneVision-7B suffers from significant degradation (e.g., dropping from 96.5\% to 84.7\% on Caltech-101), indicating that intrinsic ``thinking'' capabilities alone are insufficient to resolve the issue.
While I-MoF enhances visual grounding by integrating multiple visual features, it still suffers from modality interference: e.g. LLaVA-1.5-13B + I-MoF drops from 93.9\% to 70.1\% under perturbation (↓23.8\%), indicating reliance on spurious textual cues. In contrast, our method maintains perturbed performance at 98.4\% (↓0.1\%). 
On the other hand, VLMClassifier, adopts vision-only fine-tuning, which leads to critical limitations--vulnerability to cross-modal interference and degradation on VQA tasks, as LLaVA-1.5-7b + VLMClassifier-1 only reaches 35.8\%/36.2\% on MM-Bench/SeedBench, notably lower than both base LLaVA and our method (73.7\%/68.4\%), highlighting that vision-centric strategies, without addressing modality alignment, are insufficient for robust multimodal understanding. 
In text-heavy tasks, our method also achieves superior perturbed performance, indicating that addressing modality interference directly, rather than merely improving representations, is key to robust multimodal reasoning. 
Overall, unlike prior methods that often trade off between unimodal and multimodal performance, our method consistently improves both, achieving \textit{\textbf{Pareto-optimality}}.

\begin{table}
\centering
\captionof{table}{Evaluation of Caltech-101 (image-heavy) and MMLU (text-heavy) across different ablation study settings across different model families (full results in \Cref{tab:unimodal_vqa_combined_ablation_study}).}
\label{tab:ablation_caltech_MMLU}
\resizebox{\linewidth}{!}{
\begin{tabular}{llcc|cc}
\toprule
\multirow{2}{*}{\textbf{Model}} & \multirow{2}{*}{\textbf{Method}} &
\multicolumn{2}{c|}{\textbf{Caltech-101}} &
\multicolumn{2}{c}{\textbf{MMLU}} \\
& & Orig & Perturbed & Orig & Perturbed \\
\midrule
\multirow{6}{*}{LLaVA-1.5-7B}
  & Vanilla             & 97.0 & 57.4 & 62.4 & 56.4 \\
  & +$D^{\text{VQA}}$   & 96.2 & 46.3 & 61.3 & 55.5 \\
  & +$D^{\text{AUG}}$   & 98.5 & 98.6 & 78.6 & 77.2 \\
  & \quad+KL            & 98.6 & 98.7 & 81.4 & 81.2 \\
  & \quad+ADV           & 98.7 & 98.5 & 81.7 & 80.8 \\
  & Ours                & \textbf{99.3} & \textbf{99.0} & \textbf{81.8} & \textbf{81.5} \\
\midrule
\multirow{6}{*}{InstructBLIP-7B}
  & Vanilla             & 90.3 & 17.5 & 50.9 & 46.2 \\
  & +$D^{\text{VQA}}$   & 92.1 & 23.1 & 49.8 & 45.2 \\
  & +$D^{\text{AUG}}$   & 99.0 & 56.1 & 75.0 & 74.8 \\
  & \quad+JS            & 98.9 & \textbf{98.4} & 78.0 & 76.6 \\
  & \quad+ADV           & 99.1 & 85.2 & 76.8 & 76.3 \\
  & Ours                & \textbf{99.2} & 98.3 & \textbf{79.0} & \textbf{78.3} \\
\bottomrule
\end{tabular}
}
\vspace{-0.8cm}
\end{table}

\begin{table*}[ht]
\centering
\captionof{table}{Evaluation of out-of-distribution (OOD) robustness under \emph{unseen perturbation types} across both image-heavy and text-heavy tasks on all evaluated datasets (Seen: Mini-ImageNet, Caltech-101, OpenBookQA, MMLU; Unseen: Food-101, ARC Challenge).}
\label{tab:ood_results_summary}
\resizebox{0.95\linewidth}{!}{
\begin{tabular}{llcc|cc|cc||cc|cc|cc}
\toprule
\multirow{2}{*}{\textbf{Model}} & \multirow{2}{*}{\textbf{Method}} &
\multicolumn{6}{c||}{\textbf{Image-heavy}} &
\multicolumn{6}{c}{\textbf{Text-heavy}} \\
& 
& \multicolumn{2}{c|}{\textbf{Mini-ImageNet}}
& \multicolumn{2}{c|}{\textbf{Caltech-101}}
& \multicolumn{2}{c||}{\textbf{Food-101}}
& \multicolumn{2}{c|}{\textbf{OpenBookQA}}
& \multicolumn{2}{c|}{\textbf{MMLU}}
& \multicolumn{2}{c}{\textbf{ARC Challenge}} \\
& 
& Orig & OCR & Orig & OCR & Orig & OCR
& Orig & Screenshot & Orig & Screenshot & Orig & Screenshot \\
\midrule
\multirow{4}{*}{LLaVA-1.5-7B}
  & Vanilla           
    & 95.3 & 88.9 & 97.0 & 92.8 & 90.2 & 83.6
    & 62.4 & 50.2 & 46.3 & 44.8 & 53.4 & 51.8 \\
  & +$D^{\text{AUG}}$ 
    & 98.2 & 98.0 & 98.5 & 98.0 & 91.3 & 83.5
    & 78.6 & 78.0 & 51.1 & 50.6 & 55.7 & 53.8 \\
  & \quad+ADV
    & 98.7 & 98.2 & 98.7 & 98.4 & 91.8 & 86.7
    & 81.7 & 81.3 & 50.6  & 51.3 & 57.9 & 57.4 \\
  & Ours              
    & \textbf{98.6} & \textbf{98.2} & \textbf{99.3} & \textbf{99.0} & \textbf{92.1} & \textbf{89.5}
    & \textbf{81.8} & \textbf{81.2} & \textbf{51.5} & \textbf{51.3} & \textbf{62.8} & \textbf{62.6} \\
\midrule
\multirow{4}{*}{InstructBLIP-7B}
  & Vanilla           
    & 92.0 & 81.4 & 90.3 & 83.6 & 67.9 & 48.0
    & 50.9 & 40.0 & 35.3 & 34.9 & 33.7 & 36.5 \\
  & +$D^{\text{AUG}}$ 
    & 98.5 & 95.7 & 99.0 & 98.4 & 55.5 & 56.5
    & 75.0 & 74.4 & 50.0 & 49.2 & 27.3 & 30.7 \\ 
  & \quad+ADV
    & 98.7 & 97.5 & 99.5 & 98.8 & 93.2 & 85.9
    & 76.8 & 76.2 & 49.3 & 49.3 & 50.1 & 44.5\\
  & Ours              
    & \textbf{98.4} & \textbf{97.3} & \textbf{99.2} & \textbf{99.0} & \textbf{95.6} & \textbf{88.1}
    & \textbf{79.0} & \textbf{77.8} & \textbf{50.2} & \textbf{49.2} & \textbf{57.6} & \textbf{57.0} \\
\midrule
\multirow{4}{*}{Qwen2.5-VL-7B}
  & Vanilla           
    & 99.3 & 99.2 & 99.1 & 99.2 & 97.1 & 96.7
    & 85.9 & 69.3 & 69.3 & 57.0 & \textbf{85.6} & 79.6 \\
  & +$D^{\text{AUG}}$ 
    & 99.6 & 99.5 & \textbf{99.7} & 99.3 & 97.1 & 96.5
    & 90.2 & 85.8 & 70.4 & 63.7 & 82.9 & 80.2 \\
  & \quad+ADV
  & 99.4 & 99.5 & 99.6 & \textbf{99.6} & 97.3 & 96.9
  & 91.7 & \textbf{91.5} & \textbf{70.4} & 69.7 & 85.0 & 81.2 \\
  & Ours              
    & \textbf{99.6} & \textbf{99.5} & 99.6 & 99.5 & \textbf{97.5} & \textbf{97.4}
    & \textbf{90.9} & 90.7 & 69.8 & \textbf{69.7} & 84.1 & \textbf{82.6} \\
\bottomrule
\end{tabular}}
\vspace{-0.3cm}
\end{table*}


\vspace{-0.4cm}
\paragraph{Ablation Studies}

To evaluate the effectiveness of each component in our framework, we conduct a comprehensive ablation study across multiple models and scales. We compare the pretrained models with the following strategies:
\textit{FFT with $D^{\text{VQA}}$} (standard finetuning on VQA data), 
\textit{FFT with $D^{\text{AUG}}$} (supervised finetuning on mixed multi-task datasets with heuristic perturbations), 
\textit{FFT+KL/JS} (adding consistency regularization on KL or JS divergence), 
\textit{FFT+ADV} (FFT  with heuristic \& adversarial perturbations), 
and \textit{Ours} (combining both perturbation-based data augmentation and consistency regularization).
~\Cref{tab:vqa_summary} presents overall VQA performance, and ~\Cref{tab:ablation_caltech_MMLU} evaluates model robustness under unimodal settings. (~\Cref{tab:unimodal_summary} reports results with all perturbations.) Together, these results show the effectiveness of our method in improving both general VQA accuracy and robustness under modality interference.
Across all model families (Qwen2.5-VL, InstructBLIP, LLaVA-1.5) and model sizes (3B/7B/13B), our method consistently achieves best overall performance, improving accuracy on both unimodal and multimodal benchmarks. For instance, it boosts overall VQA accuracy (e.g., +14.9\% on InstructBLIP-7B), but also enhances robustness to modality interference--improve the performance  under perturbations by over 50\% on image-heavy tasks(e.g. 17.5\% $\rightarrow$ 98.3\% with InstructBLIP-7B on Caltech101). We also extend evaluation from MCQA to free-form QA and additional diagnostic experiments, including a no-mask ablation, which exhibit consistent robustness trends (see \Cref{sec:freeformVQA,apx:ablation_no_mask}).

We observe consistent improvements across both unimodal and multimodal tasks when moving from \textit{FFT w/ $D^{\text{VQA}}$} to \textit{FFT w/ $D^{\text{AUG}}$}, highlighting the importance of incorporating modality-specific supervision and heuristic perturbations.  
Building upon this, adding consistency regularization yields further gains by stabilizing model predictions under controlled perturbations on $X_I$ or $X_T$. Both KL and JS objectives lead to similar improvements, suggesting that the model equally benefits from all heuristic perturbations regardless of anchor choice. 
To further validate generalization, we introduce two real-world out-of-distribution perturbations at test time: (i) noisy OCR snippets sampled from FUNSD~\citep{funsd} as irrelevant text into image-heavy tasks; and (ii) unrelated UI screenshots from RICO~\citep{rico} as distractor images in text-heavy tasks. Beyond existing datasets, we further evaluate on Food-101~\citep{food101} (unseen image-heavy dataset) and ARC-Challenge~\citep{arcchallenge} (unseen text-heavy dataset), neither of which is used during training. 
As shown in~\Cref{tab:ood_results_summary}, our method significantly improves robustness under these unseen perturbations with consistent gains (e.g., on InstructBLIP-7B, 83.6\% $\rightarrow$ 99.0\% under OCR noise on Caltech-101). Crucially, results on new datasets confirm our method's strong OOD generalization. Under misleading text (seen perturbation) and OCR-style noise (unseen perturbation), vanilla models suffer severe degradation, while our method preserves accuracy close to clean-baseline performance. In contrast, heuristic data augmentation alone fails to generalize to unseen noise, and may even degrade performance (e.g., on InstructBLIP-7B, heuristic augmentation degrades performance on Food-101 under clean inputs (67.9\% vs. 55.5\%).
These results demonstrate that \textbf{the modest overhead of adversarial training} (see \Cref{Apx:Implementation Details}) \textbf{yields substantial gains in out-of-domain generalization} (full results in \Cref{tab:ood-results-new-perturbations} and \Cref{tab:ood-results-new-datasets}).

\vspace{-0.2cm}\section{Conclusion}

In this paper, we identify and formalize modality interference as a concrete manifestation of the broader cross-modality competency problem in Multimodal Large Language Models—namely, the inability to distinguish task-relevant from irrelevant modality signals. Through a designed perturbation-based causal evaluation experiment, we demonstrate that even state-of-the-art MLLMs systematically exhibit degraded performance under irrelevant but misleading inputs, revealing a fundamental vulnerability in their inference-time reasoning. 
To mitigate this issue, we propose a robust fine-tuning strategy that combines modality-specific data augmentation, consistency regularization, and adversarial perturbation in the embedding space. These designs explicitly constrain the model to produce stable outputs under spurious modality shifts, thereby reducing reliance on non-causal correlations and improving robustness. Extensive experiments across diverse architectures, scales, and task regimes confirm that our approach consistently improves both unimodal reasoning and multimodal generalization, achieving Pareto-optimal performance. 

\section{Impact Statement}

This paper presents work whose goal is to advance the field of machine learning by improving the understanding and robustness of multimodal large language models. The proposed analyses and training methods aim to make model behavior more reliable under modality-specific perturbations.

The societal impacts of this work are consistent with those commonly associated with advances in machine learning research, and we do not anticipate any unique ethical concerns that require special discussion.

\bibliography{sections/icml2026_conference}
\bibliographystyle{styles/icml2026}

\clearpage
\appendix
\section{Appendix Summary}

This appendix provides comprehensive supplementary materials and discussion to support the main findings of our paper on diagnosing and mitigating modality interference in MLLMs. We organize the appendix into several sections:

\textbf{Finetuning Strategies} (\Cref{Apx:Finetuning Strategy}): We elaborate on our design choice to freeze the Q-Former in InstructBLIP-based models. This decision is motivated by the need to retain strong visual representations while avoiding overfitting to perturbed or misleading multimodal inputs. (~\Cref{tab:dataset_sizes} records the size for each dataset)

\textbf{Detailed Experimental Results} (\Cref{Apx:experiments}): This section includes three key tables—~\Cref{tab:unimodal_summary}, ~\Cref{tab:unimodal_vqa_combined_ablation_study} and ~\Cref{tab:vqa_summary_extra_models}—which report model performance on unimodal and multimodal tasks under various perturbation settings and ablation conditions(additional models included). ~\Cref{tab:modality_interference_vinilla_models} records the performance of different vanilla MLLMs under modality interference across modality-heavy datasets. We also include radar plots (~\Cref{fig:radar_comparison}) that visualize task-wise robustness across different MLLMs. We provided the detailed experimental results on Qwen2.5-VL-7b~\citep{qwen25vl} and InstructBLIP-Vicuna-13b~\citep{instructblip} and make further discussion on the selection of specific consistency loss. In ~\Cref{tab:ood-results-new-datasets}, ~\Cref{tab:ood-results-new-perturbations}, we examine the generalization benefits of adversarial training by evaluating robustness under two types of out-of-distribution (OOD) perturbations: real-world OCR noise (from FUNSD~\citep{funsd}) and unrelated screenshots (from RICO~\citep{rico}) and two new datasets: Food-101~\citep{food101} and ARC Challenge~\citep{arcchallenge}. In ~\Cref{tab:cot-results}, we assess the impact of Chain-of-Thought prompting in mitigating modality interference, comparing its effectiveness against our method and standard baselines across both visual and textual modalities. In ~\Cref{tab:textvqa}, we report results on the free-form generative VQA benchmark TextVQA~\citep{textvqa}, highlighting our method’s generalizability beyond multiple-choice formats. In ~\Cref{tab:ablation_no_mask}, we make another ablation study without modality-specific masking. 

\textbf{Hyperparameter Settings} (\Cref{Apx:Parameter Setting}): We present full training configurations used in our experiments, including optimization strategies, perturbation settings, and sampling ratios for different task types. This section enables reproducibility and highlights the computational efficiency of our proposed training scheme. We provide parameter analysis on iterations of adversarial training in ~\Cref{fig:llava_pgd_compare}. 

\textbf{Compute Resource Details} (\Cref{Apx:Implementation Details}): We document hardware specifications, training durations, and resource costs for models of different scales. These details contextualize the feasibility of our approach in academic environments.

\textbf{Limitations} (\Cref{Apx:Limitations}): We discuss the granularity of our current modality interference analysis, the selections of perturbations, and propose directions for more fine-grained future studies.

\textbf{LLM Use} (\Cref{Apx:LLM_use}): Finally, we clarify that LLMs were only used to polish the writing of this paper.  

Together, these sections provide a complete view of our technical contributions, empirical findings, and responsible research considerations.

\begin{table*}[t]
\centering
\caption{Unimodal ability evaluation on image-heavy and text-heavy tasks under perturbation. Left: Mini-ImageNet and Caltech-101; Right: OpenBookQA and MMLU. UF = Unrelated Facts, MD = Misleading Descriptions, RP = Random Pixels, RI = Real Image, FB \& FW = Full Black/White Canvas. The best accuracy is marked in bold.}
\resizebox{\textwidth}{!}{
\begin{tabular}{l|l|ccc|ccc|cccc|cccc}
\toprule
\multirow{2}{*}{\textbf{Model}} & \multirow{2}{*}{\textbf{Method}} &
\multicolumn{3}{c|}{\textbf{Mini-ImageNet}} &
\multicolumn{3}{c|}{\textbf{Caltech-101}} &
\multicolumn{4}{c|}{\textbf{OpenBookQA}} &
\multicolumn{4}{c}{\textbf{MMLU}} \\
& & Orig & UF & MD & Orig & UF & MD & RP & RI & FB & FW & RP & RI & FB & FW \\
\midrule
\multirow{9}{*}{\shortstack{\textbf{Qwen2.5-VL-3B}\\\scriptsize\cite{qwen25vl}}}
& Vanilla                   & 98.9 & 98.5 & 94.9 & 98.8 & 99.0 & 94.4 & 79.9 & 74.6 & 80.0 & 79.7 & 63.5 & 61.1 & 64.0 & 63.6 \\
& FFT with $D^{\text{VQA}}$ & 98.8 & 98.5 & 95.3 & 98.8 & 98.1 & 94.3 & 80.7 & 74.3 & 80.1 & 80.2 & 63.0 & 61.7 & 63.0 & 63.3 \\
& FFT with $D^{\text{AUG}}$ & 98.8 & 98.8 & 98.6 & 99.6 & 99.3 & 99.6 & \textbf{87.1} & \textbf{86.7} & \textbf{87.4} & 87.2 & 64.8 & 63.9 & 64.8 & 64.7 \\
& \quad + KL                & 98.9 & 98.7 & 99.1 & 99.6 & 99.5 & \textbf{99.7} & 87.1 & 86.2 & 86.7 & 86.8 & \textbf{66.0} & \textbf{65.5} & \textbf{65.9} & \textbf{65.9} \\
& \quad + JS                & 99.1 & 98.8 & 98.3 & 99.6 & 99.4 & 98.0 & 85.0 & 84.2 & 85.1 & 85.1 & 65.6 & 65.1 & 65.5 & 65.5 \\
& \quad + RG ($\sigma{=}0.05$) & 99.0 & 99.1 & 98.9 & 99.5 & 99.5 & 99.2 & 86.4 & 86.9 & 86.9 & \textbf{87.2} & 64.6 & 64.3 & 64.6 & 64.8 \\
& \quad + ADV               & 99.3 & \textbf{99.3} & 99.1 & 99.5 & 99.4 & 99.5 & 86.6 & 85.8 & 86.8 & 86.6 & 65.3 & 64.0 & 65.4 & 65.3 \\
& Ours                      & \textbf{99.3} & 99.2 & \textbf{99.2} & \textbf{99.7} & \textbf{99.7} & 99.5 & 86.7 & 86.4 & 86.6 & 86.6 & 64.8 & 64.5 & 65.0 & 65.1 \\
\midrule
\multirow{8}{*}{\shortstack{\textbf{Qwen2.5-VL-7B}\\\scriptsize\cite{qwen25vl}}}
& Vanilla & 99.3 & 99.3 & 96.3 & 99.1 & 98.9 & 97.2 & 85.9 & 77.5 & 85.8 & 86.0 & 69.3 & 63.7 & 68.9 & 68.9 \\
& FFT with $D^{\text{VQA}}$ & 99.2 & 99.3 & 96.0 & 99.5 & 99.5 & 95.7 & 86.3 & 82.3 & 86.5 & 86.3 & 69.2 & 67.4 & 69.4 & 69.3 \\
& FFT with $D^{\text{AUG}}$ & 99.6 & 99.5 & 99.4 & 99.7 & \textbf{99.7} & 99.5 & 90.2 & 90.2 & 90.3 & 90.3 & 70.4 & 69.9 & 70.4 & 70.3 \\
& \quad + KL & 99.3 & 99.3 & 99.1 & 99.6 & 99.6 & 99.6 & \textbf{92.0} & 91.7 & 92.2 & 92.1 & 71.2 & 70.7 & 71.0 & 71.0 \\
& \quad + JS & 99.5 & 99.4 & 99.4 & \textbf{99.7} & 99.3 & 99.6 & 91.6 & 92.1 & \textbf{92.2} & \textbf{92.1} & \textbf{71.5} & 69.9 & \textbf{71.5} & \textbf{71.6} \\
& \quad + RG ($\sigma{=}0.05$) & 99.4 & 99.4 & 99.2 & 99.6 & 99.3 & 99.4 & 89.1 & 87.9 & 89.1 & 89.1 & 66.7 & 65.6 & 66.5 & 66.5 \\
& \quad + ADV & 99.4 & 99.3 & 99.2 & 99.6 & 99.5 & 99.6 & 91.7 & \textbf{92.2} & 91.8 & 91.8 & 70.4 & \textbf{70.0} & 70.4 & 70.4 \\
& Ours & \textbf{99.6} & \textbf{99.5} & \textbf{99.5} & 99.6 & 99.6 & \textbf{99.7} & 90.9 & 89.7 & 90.8 & 91.0 & 69.8 & 68.2 & 69.8 & 70.0 \\
\midrule
\multirow{9}{*}{\shortstack{\textbf{LLaVA-1.5-7B}\\\scriptsize\cite{llava15}}}
& Vanilla & 95.3 & 93.4 & 43.5 & 97.0 & 95.9 & 57.4 & 62.4 & 56.4 & 62.5 & 63.4 & 46.3 & 45.2 & 45.9 & 45.8 \\
& FFT with $D^{\text{VQA}}$ & 94.3 & 92.7 & 41.5 & 96.2 & 94.0 & 46.3 & 61.3 & 55.5 & 62.0 & 62.9 & 46.8 & 45.6 & 47.5 & 47.7 \\
& FFT with $D^{\text{AUG}}$ & 98.2 & 98.2 & 98.1 & 98.5 & 98.6 & 99.0 & 78.6 & 77.2 & 78.7 & 78.4 & 51.1 & 50.7 & 51.1 & 51.3 \\
& \quad + KL & \textbf{99.1} & \textbf{99.0} & 98.9 & 98.6 & 98.7 & 98.8 & 81.4 & 81.3 & 81.4 & 81.2 & 52.0 & 51.8 & 52.0 & 52.2 \\
& \quad + JS & 98.7 & 98.8 & \textbf{99.0} & 99.1 & \textbf{99.0} & 99.2 & 81.6 & \textbf{81.6} & 81.7 & 81.5 & \textbf{51.6} & \textbf{51.8} & \textbf{52.4} & \textbf{52.4} \\
& \quad + RG ($\sigma{=}0.05$) & 98.4 & 98.4 & 98.5 & 98.9 & 98.9 & 99.1 & 80.5 & 79.8 & 79.9 & 80.3 & 49.5 & 49.5 & 49.9 & 49.6 \\
& \quad + ADV & 98.7 & 98.7 & 98.5 & 98.7 & 98.5 & 98.8 & 81.7 & 81.0 & 81.4 & 80.8 & 50.6 & 50.3 & 50.9 & 50.7 \\
& Ours & 98.6 & 98.4 & 98.7 & \textbf{99.3} & 98.9 & \textbf{99.3} & \textbf{81.8} & 81.0 & \textbf{81.7} & \textbf{81.7} & 51.5 & 50.9 & 51.5 & 51.4 \\
\midrule
\multirow{9}{*}{\shortstack{\textbf{LLaVA-1.5-13B}\\\scriptsize\cite{llava15}}}
& Vanilla        & 95.6 & 94.1 & 73.0 & 97.9 & 97.1 & 77.4 & 65.9 & 63.8 & 68.0 & 69.1 & 51.8 & 50.8 & 52.7 & 52.7 \\
& FFT with $D^{\text{VQA}}$ & 94.6 & 93.9 & 72.0 & 97.8 & 96.5 & 80.2 & 67.5 & 64.2 & 69.1 & 69.3 & 52.4 & 52.2 & 53.1 & 53.3 \\
& FFT with $D^{\text{AUG}}$ & 98.1 & 96.8 & 98.4 & 96.7 & 96.9 & 97.0 & 81.0 & 78.7 & 81.1 & 81.3 & 52.1 & 51.7 & 51.8 & 51.6 \\
& \quad + KL       & 98.3 & 98.0 & 98.6 & 98.8 & 98.5 & 98.9 & 83.0 & 82.6 & \textbf{83.3} & 83.0 & 55.7 & 55.1 & 55.6 & 55.6 \\
& \quad + JS       & 98.3 & 98.1 & 98.0 & 98.7 & 98.4 & 98.7 & \textbf{83.1} & 81.5 & 83.1 & 83.1 & 56.7 & \textbf{56.2} & 56.6 & 56.5 \\
& \quad + RG ($\sigma{=}0.05$) & 98.5 & 98.0s & 98.1 & 98.9 & 98.5 & 98.9 & 83.5 & 82.5 & 83.1 & 82.8 & 55.4 & 55.3 & 55.7 & 55.5\\
& \quad + ADV      & \textbf{98.7} & 98.2 & 98.6 & 99.0 & 98.6 & 99.0 & 82.2 & \textbf{82.6} & 82.6 & 82.8 & 55.6 & 55.4 & 55.6 & 55.5 \\
& Ours           & 98.5 & \textbf{98.4} & \textbf{98.7} & \textbf{99.2} & \textbf{98.6} & \textbf{99.2} & 83.0 & 82.1 & 82.7 & \textbf{83.1} & \textbf{56.7} & 55.8 & \textbf{56.7} & \textbf{56.7} \\
\midrule
\multirow{9}{*}{\shortstack{\textbf{InstructBLIP-7B}\\\scriptsize\cite{instructblip}}}
& Vanilla        & 92.0 & 87.1 & 13.6 & 90.3 & 90.2 & 17.5 & 50.8 & 46.2 & 50.9 & 50.7 & 35.3 & 35.8 & 35.2 & 35.7 \\
& FFT with $D^{\text{VQA}}$ & 95.6 & 86.6 & 16.3 & 98.3 & 91.0 & 23.1 & 49.8 & 45.2 & 49.5 & 50.7 & 40.9 & 40.2 & 41.0 & 41.6 \\
& FFT with $D^{\text{AUG}}$ & 98.5 & 98.0 & 38.2 & 99.0 & 98.7 & 56.1 & 75.0 & 74.9 & 74.8 & 75.8 & 50.0 & 49.7 & 50.0 & 50.0 \\
& \quad + KL       & 98.7 & 98.1 & \textbf{98.3} & 99.5 & \textbf{99.0} & \textbf{99.6} & 76.9 & 77.0 & 76.9 & 77.3 & \textbf{51.3} & \textbf{50.6} & \textbf{51.3} & \textbf{51.5} \\ 
& \quad + JS       & 98.5 & 97.7 & 98.5 & 98.9 & 98.4 & 98.9 & 78.0 & 76.6 & 77.7 & 78.0 & 50.7 & 50.1 & 50.7 & 50.8 \\ 
& \quad + RG ($\sigma{=}0.05$)     &  98.9 & 97.2 & 72.5 & 99.1 & 99.0 & 82.2 & 75.2 & 72.6 & 76.0 & 76.9 & 48.3 & 47.6 & 48.9 & 49.1 \\
& \quad + ADV      & \textbf{98.7} & \textbf{98.5} & 32.2 & \textbf{99.5} & 98.9 & 49.2 & 76.8 & 76.8 & 76.5 & 76.3 & 49.3 & 48.4 & 49.5 & 49.4 \\
& Ours & 98.4 & 97.9 & 98.0 & 99.2 & 98.3 & 99.0 & \textbf{79.0} & \textbf{77.3} & \textbf{79.3} & \textbf{79.0} & 50.2 & 49.7 &	50.3 & 50.2 \\
\midrule
\multirow{9}{*}{\shortstack{\textbf{InstructBLIP-13B}\\\scriptsize\cite{instructblip}}}
& Vanilla & 95.6 & 94.1 & 73.0 & 97.9 & 97.1 & 77.4 & 65.9 & 63.8 & 68.0 & 69.1 & 51.8 & 50.8 & 52.7 & 52.7 \\
& FFT with $D^{\text{VQA}}$ & 95.6 & 85.8 & 8.0 & 97.0 & 87.5 & 11.6 & 58.6 & 55.4 & 59.7 & 60.6 & 43.7 & 42.8 & 43.6 & 44.1 \\
& FFT with $D^{\text{AUG}}$ & 98.4 & 98.2 & 9.3 & 99.2 & 98.8 & 13.8 & 82.0 & 80.4 & 81.2 & 81.2 & 52.1 & 51.3 & 52.4 & 53.0 \\
& \quad + KL & 98.5 & \textbf{98.3} & 98.7 & 99.1 & 99.2 & 99.5 & 82.5 & 81.4 & 82.1 & 82.9 & \textbf{53.4} & \textbf{52.5} & \textbf{53.4} & 53.4 \\
& \quad + JS & 98.7 & 98.1 & \textbf{98.9} & 99.3 & 99.2 & \textbf{99.5} & \textbf{83.5} & \textbf{83.1} & 83.1 & \textbf{83.3} & 52.8 & 52.2 & 53.2 & 53.3 \\
& \quad + RG ($\sigma{=}0.05$) & 98.4 & 97.8 & 87.0 & \textbf{99.4} & \textbf{99.3} & 94.4 & 80.0 & 76.6 & 79.6 & 80.4 & 50.9 & 50.0 & 51.4 & 51.8 \\
& \quad + ADV & 98.6 & 98.0 & 80.9 & 98.7 & 98.6 & 99.1 & 79.8 & 79.0 & 80.9 & 80.9 & 51.3 & 50.7 & 51.4 & 52.4 \\
& Ours & \textbf{98.7} & 97.9 & 98.0 & 98.7 & 98.7 & 98.8 & 83.2 & 81.2 & \textbf{83.8} & 83.0 & 52.2 & 51.6 & 52.3 & \textbf{53.4} \\
\bottomrule
\end{tabular}
}
\label{tab:unimodal_summary}
\end{table*}

\begin{table*}[t]
\centering
\caption{Evaluation of unimodal and multimodal tasks across different ablation study settings. For unimodal datasets, we report accuracy on the original setting (Orig) and the worst-performing perturbation (Perturbed). For VQA datasets, we report accuracy on the original setting. The best accuracy is marked in bold.}
\vspace{0.6em}
\label{tab:unimodal_vqa_combined_ablation_study}
\resizebox{\textwidth}{!}{
\begin{tabular}{llcc|cc|cc|cc|c}
\toprule
\multirow{2}{*}{\textbf{Model}} & \multirow{2}{*}{\textbf{Method}} &
\multicolumn{2}{c}{\textbf{Mini-ImageNet}} &
\multicolumn{2}{c}{\textbf{Caltech-101}} &
\multicolumn{2}{c}{\textbf{OpenBookQA}} &
\multicolumn{2}{c}{\textbf{MMLU}} &
\multicolumn{1}{c}{\textbf{VQA Overall}}\\
& & Orig & Perturbed & Orig & Perturbed & Orig & Perturbed & Orig & Perturbed & Accuracy \\
\midrule
\rowcolor{gray!40}
\multicolumn{11}{c}{\textbf{\textit{3B Multimodal Models}}} \\
\midrule
\multirow{9}{*}{\shortstack{\textbf{Qwen2.5-VL-3B}\\\scriptsize\cite{qwen25vl}}}
 & Vanilla & 98.9 & 94.9 & 98.8 & 94.4 & 79.9 & 74.6 & 63.5 & 61.1 & 72.5 \\
 & FFT with $D^{\text{VQA}}$ & 98.8 & 95.3 & 98.8 & 94.3 & 80.7 & 74.3 & 63.0 & 61.7 & 75.5 \\
 & FFT with $D^{\text{AUG}}$ & 98.8 & 98.6 & 99.6 & 99.3 & 87.1 & 86.2 & 64.8 & 63.9 & 76.2 \\
 & \quad + KL                & 98.9 & 98.7 & 99.6 & 99.5 & \textbf{87.1} & \textbf{86.7} & \textbf{66.0} & \textbf{65.5} & 75.9 \\
 & \quad + JS                & 99.1 & 98.3 & 99.6 & 98.0 & 85.0 & 84.2 & 65.6 & 65.1 & 76.2 \\
 & \quad + RG                & 99.0 & 98.9 & 99.5 & 99.2 & 86.4 & 86.4 & 64.6 & 64.3 & 76.0 \\
 & \quad + ADV               & 99.3 & 99.1 & 99.5 & 99.4 & 86.6 & 85.8 & 65.3 & 64 & 76.1 \\
 & Ours                      & \textbf{99.3} & \textbf{99.2} & \textbf{99.7} & \textbf{99.5} & 86.7 & 86.6 & 64.8 & 64.5 & \textbf{76.4} \\
\midrule
\rowcolor{gray!40}
\multicolumn{11}{c}{\textbf{\textit{7B Multimodal Models}}} \\
\midrule
\multirow{9}{*}{\shortstack{\textbf{LLaVA-1.5-7B}\\\scriptsize\cite{llava15}}}
 & Vanilla                   & 95.3 & 43.5 & 97.0 & 57.4 & 62.4 & 56.4 & 46.3 & 45.2 & 63.8 \\
 & FFT with $D^{\text{VQA}}$ & 94.3 & 41.5 & 96.2 & 46.3 & 61.3 & 55.5 & 46.8 & 45.6 & 64.0 \\
 & FFT with $D^{\text{AUG}}$ & 98.2 & 98.1 & 98.5 & 98.6 & 78.6 & 77.2 & 51.1 & 50.7 & 65.6 \\
 & \quad + KL                & \textbf{99.1} & \textbf{99.0} & 98.6 & 98.7 & 81.4 & 81.2 & 51.1 & 50.7 & 66.3 \\
 & \quad + JS                & 98.7 & 98.8 & 99.1 & 99.0 & 81.6 & 81.5 & \textbf{52.0} & \textbf{51.8} & 65.6 \\
 & \quad + RG                & 98.4 & 98.4 & 98.9 & 98.9 & 80.5 & 79.8 & 49.5 & 49.5 & 65.7 \\
 & \quad + ADV               & 98.7 & 98.5 & 98.7 & 98.5 & 81.7 & 80.8 & 50.6 & 50.3 & 65.7 \\
 & Ours                      & 98.6 & 98.6 & \textbf{99.3} & \textbf{99.0} & \textbf{81.8} & \textbf{81.5} & 51.5 & 51.0 & \textbf{66.8} \\
\midrule
\multirow{9}{*}{\shortstack{\textbf{InstructBLIP-7B}\\\scriptsize\cite{instructblip}}}
 & Vanilla                   & 92.0 & 13.6 & 90.3 & 17.5 & 50.9 & 46.2 & 35.3 & 35.2 & 56.4 \\
 & FFT with $D^{\text{VQA}}$ & 95.6 & 16.3 & 98.3 & 23.1 & 49.8 & 45.2 & 40.9 & 40.2 & 60.4 \\
 & FFT with $D^{\text{AUG}}$ & 98.5 & 38.2 & 99.0 & 56.1 & 75.0 & 74.8 & 50.0 & 49.7 & 63.9 \\
 & \quad + KL                & 98.7 & \textbf{98.1} & 99.5 & \textbf{99.0} & 76.9 & 76.9 & \textbf{51.3} & \textbf{50.6} & 63.9 \\
 & \quad + JS                & 98.5 & 97.7 & 98.9 & 98.4 & 78.0 & 76.6 & 50.7 & 50.1 & 64.1 \\
 & \quad + RG                & \textbf{98.9} & 72.5 & 99.1 & 82.2 & 76.0 & 72.6 & 48.3 & 47.6 & 64.2 \\
 & \quad + ADV               & 98.7 & 72.2 & \textbf{99.5} & 85.2 & 76.8 & 76.3 & 49.3 & 48.4 & 64.0 \\
 & Ours                      & 98.4 & 98.0 & 99.2 & 98.3 & \textbf{79.0} & \textbf{78.3} & 50.2 & 49.7 & \textbf{64.8} \\
\midrule
\multirow{9}{*}{\shortstack{\textbf{Qwen2.5-VL-7B}\\\scriptsize\cite{instructblip}}}
 & Vanilla                   & 99.3 & 96.3 & 99.1 & 97.2 & 85.9 & 77.5 & 69.3 & 63.7 & 80.3 \\
 & FFT with $D^{\text{VQA}}$ & 99.2 & 96.0 & 99.5 & 95.7 & 86.3 & 82.3 & 69.2 & 67.4 & 79.5 \\
 & FFT with $D^{\text{AUG}}$ & 99.6 & 99.4 & \textbf{99.7} & 99.5 & 90.2 & 90.2 & 70.4 & 69.9 & 79.9 \\
 & \quad + KL                & 99.3 & 99.1 & 99.6 & 99.6 & \textbf{92.0} & 91.7 & 71.2 & \textbf{70.7} & 80.6 \\
 & \quad + JS                & 99.5 & 99.4 & 99.7 & 99.3 & 91.6 & \textbf{92.1} & \textbf{71.5} & 69.9 & 80.3 \\
 & \quad + RG                & 99.4 & 99.3 & 99.7 & 99.5 & 91.7 & 91.8 & 70.4 & 69.9 & 78.0 \\
 & \quad + ADV               & 99.4 & 99.2 & 99.6 & 99.5 & 91.7 & 91.8 & 70.4 & 70.0 & 79.9 \\
 & Ours                      & \textbf{99.6} & \textbf{99.5} & 99.6 & \textbf{99.7} & 90.9 & 89.7 & 69.8 & 68.2 & \textbf{80.9} \\
\midrule
\rowcolor{gray!40}
\multicolumn{11}{c}{\textbf{\textit{13B Multimodal Models}}} \\
\midrule
\multirow{9}{*}{\shortstack{\textbf{LLaVA-1.5-13B}\\\scriptsize\cite{llava15}}}
 & Vanilla                   & 95.6 & 73.0 & 97.9 & 77.4 & 65.9 & 63.8 & 51.8 & 50.8 & 66.2 \\
 & FFT with $D^{\text{VQA}}$ & 94.6 & 72.0 & 97.8 & 80.2 & 67.5 & 64.2 & 52.4 & 52.2 & 65.8 \\
 & FFT with $D^{\text{AUG}}$ & 98.1 & 96.8 & 96.7 & 96.9 & 81.0 & 78.7 & 52.1 & 51.6 & 67.1 \\
 & \quad + KL                & 98.3 & 98.0 & 98.8 & 98.5 & 83.0 & \textbf{82.6} & 55.7 & 55.1 & 68.0 \\
 & \quad + JS                & 98.3 & 98.1 & 98.7 & 98.4 & 83.1 & 81.5 & 56.7 & \textbf{56.2} & \textbf{68.6} \\
 & \quad + RG                & 98.5 & 98.0 & 98.9 & 98.5 & \textbf{83.5} & 82.5 & 55.4 & 55.3 & 66.9 \\
 & \quad + ADV               & \textbf{98.7} & 98.2 & 99.0 & 98.6 & 82.2 & 82.5 & 55.6 & 55.4 & 67.5 \\
 & Ours                      & 98.5 & \textbf{98.4} & \textbf{99.2} & \textbf{98.7} & 83.0 & 82.1 & \textbf{56.7} & 56.0 & 68.4 \\
\midrule
\multirow{9}{*}{\shortstack{\textbf{InstructBLIP-13B}\\\scriptsize\cite{instructblip}}}
 & Vanilla                   & 95.6 & 73.0 & 97.9 & 77.4 & 65.9 & 63.8 & 51.8 & 50.8 & 65.8\\
 & FFT with $D^{\text{VQA}}$ & 95.6 & 8.0  & 97.0 & 11.6 & 58.6 & 55.4 & 43.7 & 42.8 & 59.4 \\
 & FFT with $D^{\text{AUG}}$ & 98.4 & 9.3  & 99.2 & 13.8 & 82.0 & 80.4 & 52.1 & 51.3 & 65.9 \\
 & \quad + KL                & 98.5 & \textbf{98.3} & 99.1 & 99.2 & 82.5 & 81.4 & \textbf{53.4} & \textbf{52.5} & 66.2 \\
 & \quad + JS                & 98.7 & 98.1 & 99.3 & 99.2 & \textbf{83.5} & \textbf{83.1} & 52.8 & 52.2 & 66.5 \\
 & \quad + RG                & 98.4 & 97.8 & \textbf{99.4} & \textbf{99.3} & 80.0 & 76.6 & 50.9 & 50.0 & 66.3 \\
 & \quad + ADV               & 98.6 & 80.9 & 98.7 & 98.6 & 79.8 & 79.0 & 51.3 & 50.7 & 66.4 \\
 & Ours                      & \textbf{98.7} & 97.9 & 98.7 & 98.7 & 83.2 & 81.2 & 52.2 & 51.6 & \textbf{66.5} \\
\bottomrule
\end{tabular}
}
\vspace{-0.3cm}
\end{table*}

\section{Finetuning Strategies}
\label{Apx:Finetuning Strategy}

In our adaptation of InstructBLIP-Vicuna-7B, we choose to freeze the Q-Former and only fine-tune the language model and the projection layer. This decision is grounded in the nature of the Q-Former as a highly task-specific visual query encoder, originally pre-trained on VQA-style datasets where fine-grained and semantically aligned image-text pairs dominate.

However, in our setting, we deliberately introduce perturbations to the input modalities (e.g., injecting unrelated or misleading text/image content), which breaks the expected alignment structure. We observe that training the Q-Former under such noisy supervision leads to unstable representations and overfitting to spurious modality correlations. In contrast, freezing the Q-Former allows us to preserve its original strong visual grounding capabilities, while letting the downstream language model learn to filter or suppress misleading signals introduced during training.

This alternative tuning strategy enhances robustness under modality interference and aligns with our overall goal of improving cross-modal competency in MLLMs under perturbed conditions.

\section{Detailed Experimental Results}
\label{Apx:experiments}
\begin{figure*}[t]
\centering
\begin{minipage}[t]{0.49\textwidth}
    \centering
    \includegraphics[width=\linewidth]{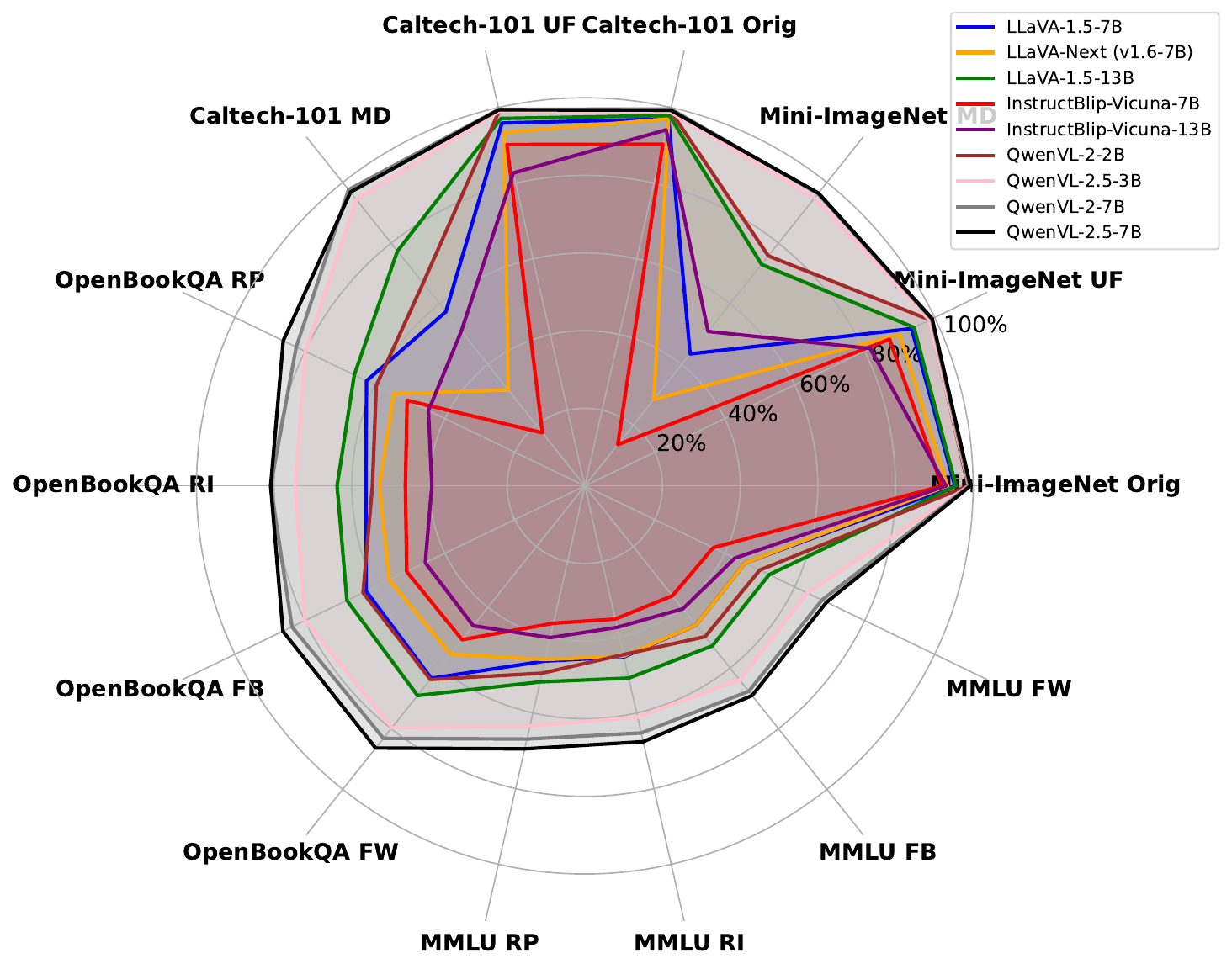}
    \vspace{1mm}
    \scriptsize (a) Pretrained MLLMs
\end{minipage}
\hfill
\begin{minipage}[t]{0.49\textwidth}
    \centering
    \includegraphics[width=\linewidth]{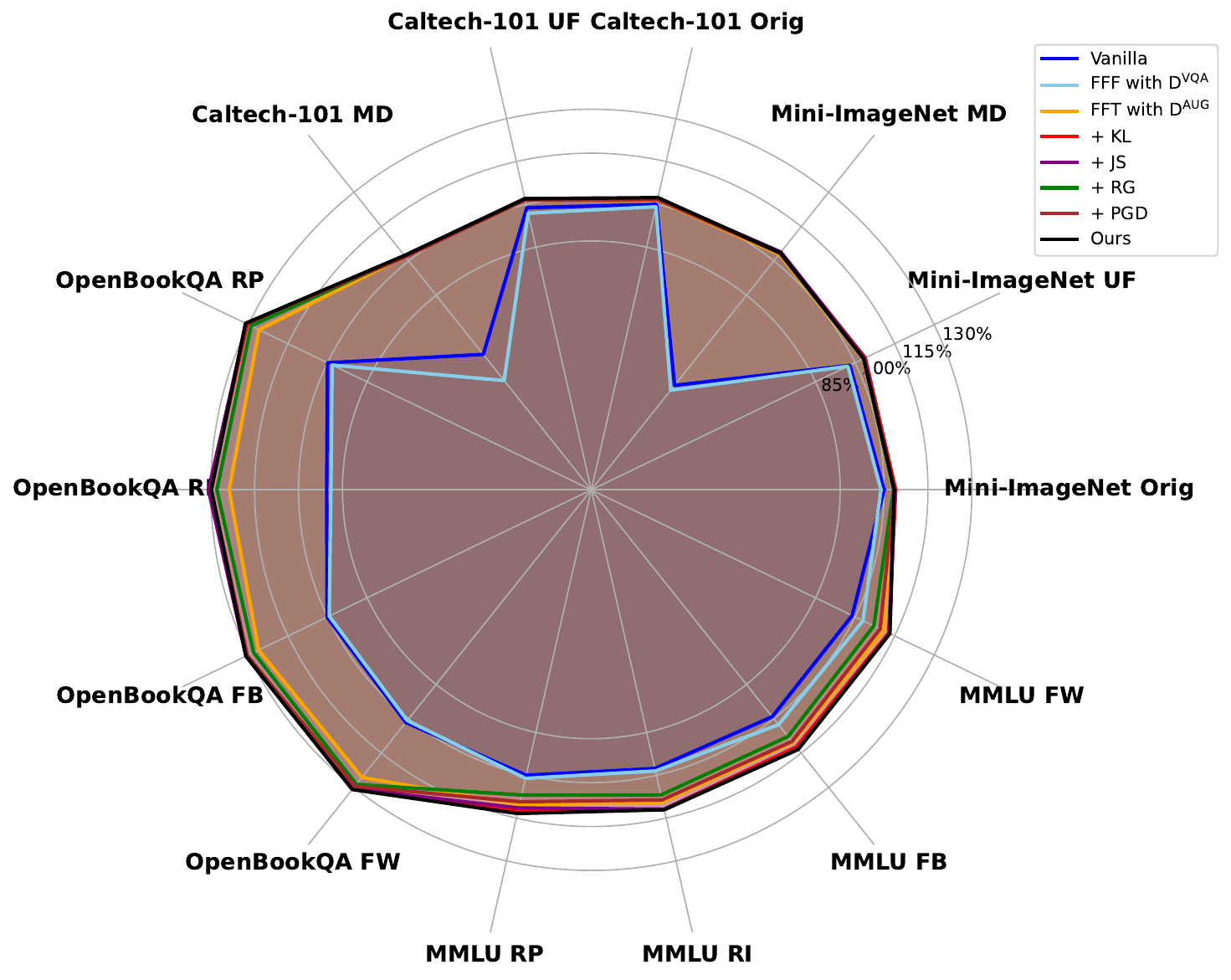}
    \vspace{1mm}
    \scriptsize (b) LLaVA-1.5-7B
\end{minipage}
\vspace{4mm}
\begin{minipage}[t]{0.49\textwidth}
    \centering
    \includegraphics[width=\linewidth]{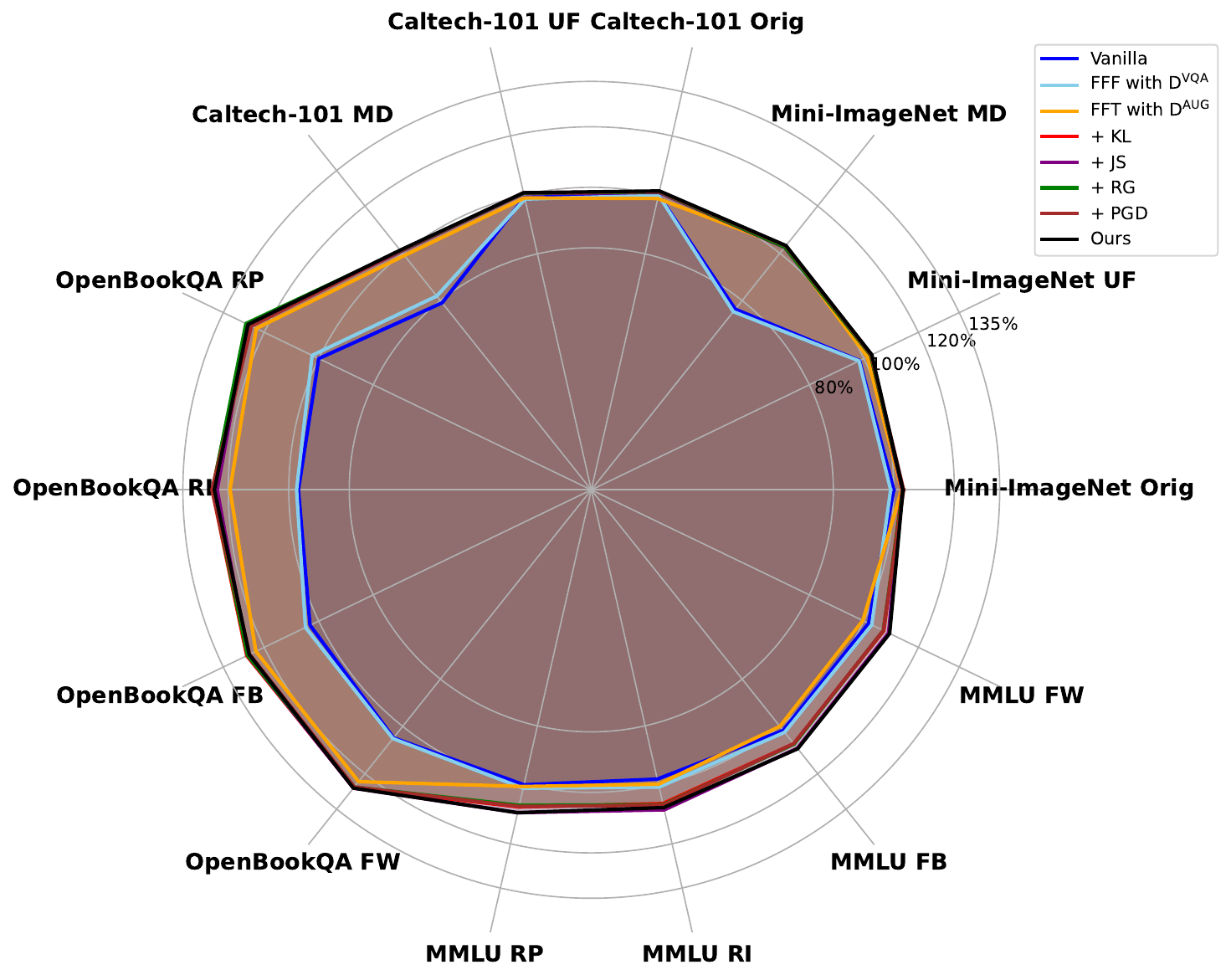}
    \vspace{1mm}
    \scriptsize (c) LLaVA-1.5-13B
\end{minipage}
\hfill
\begin{minipage}[t]{0.49\textwidth}
    \centering
    \includegraphics[width=\linewidth]{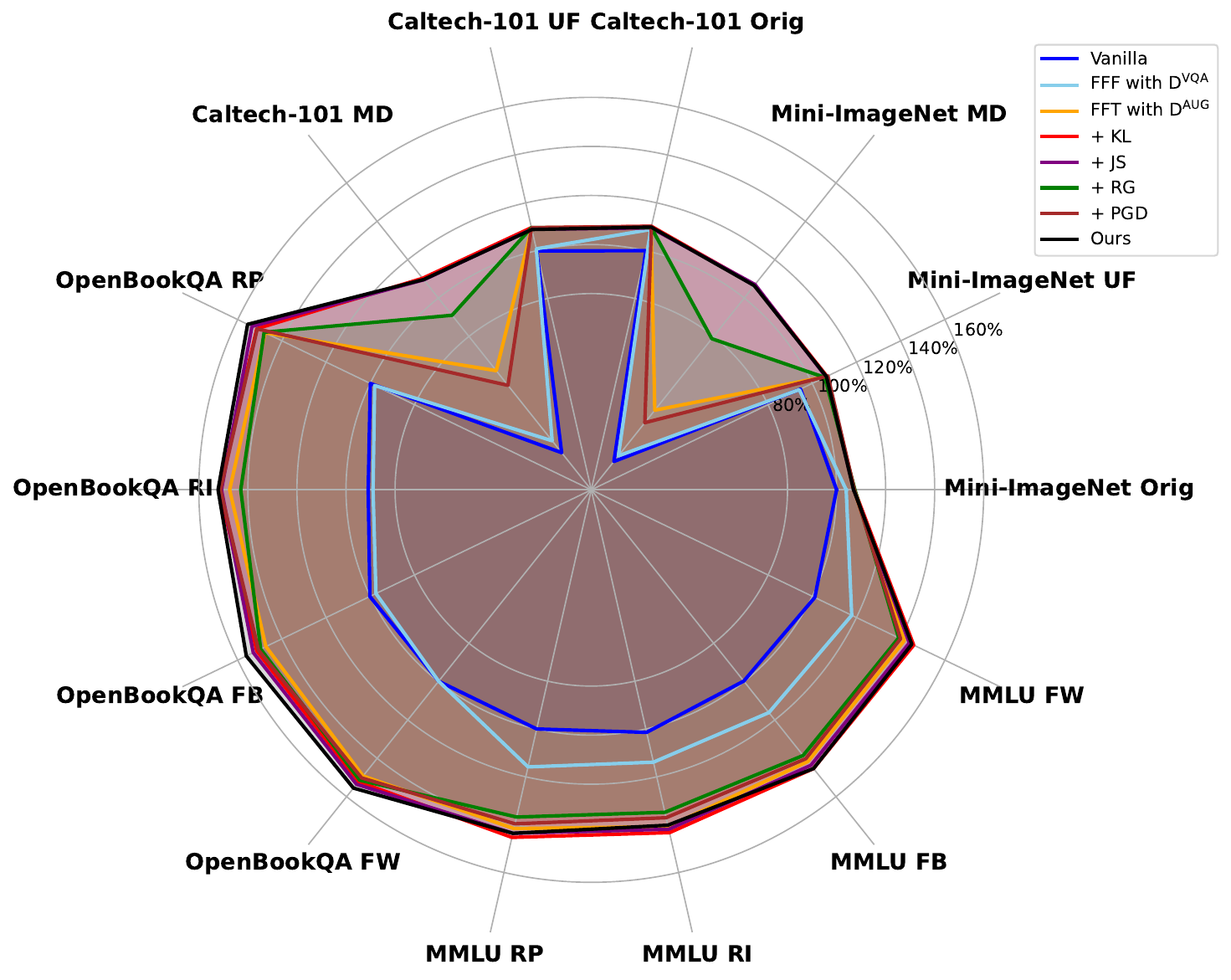}
    \vspace{1mm}
    \scriptsize (d) InstructBLIP-7B
\end{minipage}
\vspace{2mm}
\caption{
Task-wise robustness under perturbation. Each radar chart shows model accuracy (\%) across Mini-ImageNet, Caltech-101 (image-heavy) and OpenBookQA, MMLU (text-heavy) under various perturbations. (a) uses raw accuracy of different pretrained MLLMs directly. (b–d) are normalized relative accuracy of each MLLMs. (We normalize each absolute accuracy into relative accuracy, which refers to absolute tested accuracy / accuracy of vanilla MLLMs in origin setting without perturbation.)
}
\label{fig:radar_comparison}
\end{figure*}

Please see ~\Cref{tab:unimodal_summary}, ~\Cref{tab:unimodal_vqa_combined_ablation_study}, ~\Cref{fig:radar_comparison}, ~\Cref{tab:modality_interference_vinilla_models}, ~\Cref{tab:vqa_summary_extra_models}, ~\Cref{tab:ood-results-new-datasets}, ~\Cref{tab:ood-results-new-perturbations}, ~\Cref{tab:cot-results} and ~\Cref{tab:textvqa} for more details. In table ~\Cref{tab:vqa_summary_extra_models}: We compare the pretrained models with the following strategies:
\textit{FFT with $D^{\text{VQA}}$} (standard finetuning on VQA data), 
\textit{FFT with $D^{\text{AUG}}$} (supervised finetuning on mixed multi-task datasets with heuristic perturbations), 
\textit{FFT+KL/JS} (adding consistency regularization on KL or JS divergence), 
\textit{FFT+RG} (injecting random Gaussian noise into token embeddings), 
\textit{FFT+ADV} (FFT  with heuristic \& adversarial perturbations), 
and \textit{Ours} (combining both perturbation-based data augmentation and consistency regularization).

\begin{table*}[t] 
\centering
\small
\caption{Detailed multimodal reasoning accuracy (\%) on multiple-choice VQA datasets across different ablation study settings with extra models: Qwen2.5-vl-7B, Instructblip-Vicuna-13B. The best accuracy is marked in bold. Overall performance is computed as a weighted average across datasets, with weights proportional to each dataset’s test size.}
\label{tab:vqa_summary_extra_models}
\resizebox{\textwidth}{!}{
\begin{tabular}{llcccc}
\toprule
\textbf{Model} & \textbf{Method} & \textbf{ScienceQA-IMG} & \textbf{MM-Bench-EN} & \textbf{Seed-Bench-IMG} & \textbf{VQA Overall} \\
\midrule
\rowcolor{gray!40}
\multicolumn{6}{c}{\textbf{\textit{Qwen2.5-VL Models~\citep{qwen25vl}}}} \\
\midrule
\multirow{9}{*}{\shortstack{\textbf{Qwen2.5-VL-3B}\\\scriptsize\cite{qwen25vl}}}
& Vanilla & 63.0 (100\%) & 81.8 (100\%) &	72.3 (100\%) & 72.5 (100\%) \\
& FFT with $D^{\text{VQA}}$ &  72.2 (↑14.6\%) & 82.0 (↑0.2\%) & 74.7 (↑3.3\%) & 75.5 (↑3.9\%) \\
& FFT with $D^{\text{AUG}}$ & 73.6 (↑16.8\%) & 81.7 (↓0.1\%) & 75.3 (↑4.1\%) & 76.2 (↑5.1\%)\\
& \quad + KL & 72.8 (↑15.6\%) & 80.9 (↓1.1\%) & 74.9 (↑3.6\%) & 75.9 (↑4.7\%) \\
& \quad + JS & 73.4 (↑16.5\%) & 82.0 (↑0.2\%) & 75.2 (↑4.0\%) & 76.2 (↑5.1\%) \\
& \quad + RG & 73.4 (↑16.5\%) & 81.8 (↑0.0\%)	& 75.0 (↑3.7\%) & 76.0 (↑4.8\%)\\
& \quad + ADV & 73.1 (↑16.0\%) & 81.7 (↓0.1\%) & 75.3 (↑4.1\%) & 76.1 (↑5.0\%) \\
& Ours & 73.8 (↑17.1\%) & 82.1 (↑0.1 \%) & 75.5 (↑4.4\%) & \cellcolor{gray!20}\textbf{76.4 (↑5.4\%)}\\
\midrule
\multirow{9}{*}{\shortstack{\textbf{Qwen2.5-VL-7B}\\\scriptsize\cite{qwen25vl}}}
& Vanilla                   & 85.1 (100\%) & 86.6 (100\%) & 77.0 (100\%) & 80.3 (100\%) \\
& FFT with $D^{\text{VQA}}$ & 81.4 (↓4.3\%) & 86.4 (↓0.2\%) & 76.8 (↓0.3\%) & 79.5 (↓1.0\%)\\
& FFT with $D^{\text{AUG}}$ & 83.2 (↓2.2\%) & 86.1 (↓0.6\%) & 77.0 (-0.0\%) & 79.9 (↓0.5\%) \\
& \quad + KL                & 86.0 (↑1.1\%) & 86.7 (↑0.1\%) & 77.2 (↑0.3\%) & 80.6 (↑0.4\%)\\
& \quad + JS                & 85.5 (↑0.5\%) & 85.9 (↓0.8\%) & 77.0 (-0.0\%) & 80.3 (-0.0\%) \\
& \quad + RG                & 81.6 (↓4.1\%) & 82.9 (↓4.2\%) & 75.5 (↓1.9\%) & 78.0 (↓2.8\%) \\
& \quad + ADV               & 85.2 (↑0.1\%) & 85.8 (↓0.9\%) & 76.6 (↓0.5\%) & 79.9 (↓0.5\%) \\
& Ours                      & 83.9 (↓1.4\%) & 86.4 (↓0.2\%) & 78.3 (↑1.7\%) & \cellcolor{gray!20}\textbf{80.9 (↑0.7\%)}\\
\midrule
\rowcolor{gray!40}
\multicolumn{6}{c}{\textbf{\textit{Instructblip-Vicuna Models~\citep{instructblip}}}} \\
\midrule
\multirow{9}{*}{\shortstack{\textbf{InstructBLIP-7B}\\\scriptsize\cite{instructblip}}}
& Vanilla & 52.1 (100\%) & 65.5 (100\%) & 54.8 (100\%) & 56.4 (100\%) \\
& FFT with $D^{\text{VQA}}$ & 61.8 (↑18.6\%)& 68.4 (↑4.4\%) & 57.6 (↑5.1\%) &	60.4 (↑7.1\%)\\
& FFT with $D^{\text{AUG}}$ & 65.4 (↑25.5\%) & 71.1 (↑8.5\%) & 61.2 (↑11.7\%) & 63.9 (↑13.3\%) \\
& \quad + KL & 65.9 (↑26.5\%) & 71.7 (↑9.5\%) & 60.9 (↑11.1\%) & 63.9 (↑13.3\%) \\
& \quad + JS & 66.9 (↑28.4\%) & 72.3 (↑10.4\%) & 60.7 (↑10.8\%) & 64.1 (↑13.7\%) \\
& \quad + RG & 62.8 (↑20.5\%) & 70.9 (↑8.2\%) & 62.5 (↑14.0\%) & 64.2 (↑13.8\%) \\
& \quad + ADV & 66.0 (↑26.7\%) & 69.0 (↑5.3\%) & 61.8 (↑12.8\%) & 64.0 (↑13.5\%) \\
& Ours & 64.0 (↑22.8\%) & 71.4 (↑9.0\%) & 63.0 (↑14.9\%) & \cellcolor{gray!20}\textbf{64.8 (↑14.9\%)} \\
\midrule
\multirow{9}{*}{\shortstack{\textbf{InstructBLIP-13B}\\\scriptsize\cite{instructblip}}}
& Vanilla                   & 65.8 (100\%) & 72.1 (100\%) & 63.8 (100\%) & 65.8 (100\%) \\
& FFT with $D^{\text{VQA}}$ & 61.7 (↓6.2\%) & 68.5 (↓5.0\%) & 56.0 (↓12.2\%) & 59.4 (↓9.7\%) \\
& FFT with $D^{\text{AUG}}$ & 66.6 (↑1.2\%) & 71.5 (↓0.8\%) & 64.0 (↑0.3\%) & 65.9 (↑0.2\%) \\
& \quad + KL                & 67.2 (↑2.2\%) & 72.6 (↑0.7\%) & 64.0 (↑0.3\%) & 66.2 (↑0.6\%) \\
& \quad + JS                & 67.9 (↑3.2\%) & 72.7 (↑0.8\%) & 64.2 (↑0.6\%) & 66.5 (↑1.1\%) \\
& \quad + RG                & 65.4 (↓0.6\%) & 73.5 (↑1.9\%) & 64.3 (↑0.7\%) & 66.3 (↑0.8\%) \\
& \quad + ADV               & 66.1 (↑0.5\%) & 74.0 (↑2.6\%) & 64.1 (↑0.5\%) & 66.4 (↑0.9\%) \\
& Ours                      & 66.2 (↑0.6\%) & 73.2 (↑1.5\%) & 64.3 (↑0.7\%) & \cellcolor{gray!20}\textbf{66.5 (↑1.1\%)} \\
\midrule
\rowcolor{gray!40}
\multicolumn{6}{c}{\textbf{\textit{LLaVA1.5 Models~\citep{llava15}}}} \\
\midrule
\multirow{9}{*}{\shortstack{\textbf{LLaVA-1.5-7B}\\\scriptsize\cite{llava15}}}
& Vanilla & 64.5 (100\%) & 64.3 (100\%) & 63.4 (100\%) & 63.8 (100\%) \\
& FFT with $D^{\text{VQA}}$ & 61.6 (↓4.5\%) & 71.4 (↑11.0\%) & 62.4 (↓1.6\%) & 64.0 (↑0.3\%) \\
& FFT with $D^{\text{AUG}}$ & 65.4 (↑1.4\%) & 71.1 (↑10.6\%) & 63.9 (↑0.8\%) & 65.6 (↑2.8\%) \\
& \quad + KL & 65.9 (↑2.2\%) & 72.5 (↑12.8\%) & 64.5 (↑1.7\%) & 66.3 (↑3.9\%) \\
& \quad + JS & 63.8 (↓1.1\%) & 73.5 (↑14.3\%) & 63.7 (↑0.5\%) & 65.6 (↑2.8\%) \\
& \quad + RG & 66.3 (↑2.8\%) & 71.8 (↑11.7\%) & 63.7 (↑0.5\%) & 65.7 (↑3.0\%) \\
& \quad + ADV & 66.7 (↑3.4\%) & 71.4 (↑11.0\%) & 63.6 (↑0.3\%) & 65.7 (↑3.0\%) \\
& Ours & 67.8 (↑5.1\%) & 73.1 (↑13.7\%) & 64.6 (↑1.9\%) & \cellcolor{gray!20}\textbf{66.8 (↑4.7\%)} \\
\midrule
\multirow{9}{*}{\shortstack{\textbf{LLaVA-1.5-13B}\\\scriptsize\cite{llava15}}}
 & Vanilla   & 65.8 (100\%) & 72.1 (100\%) & 64.5 (100\%) & 66.2 (100\%) \\
 & FFT with $D^{\text{VQA}}$ & 60.8 (↓7.6\%) & 73.6 (↑2.1\%) & 64.9 (↑0.6\%) & 65.8 (↓0.6\%) \\
& FFT with $D^{\text{AUG}}$& 63.5 (↓3.5\%) & 75.0 (↑4.0\%) & 65.7 (↑1.9\%) & 67.1 (↑1.4\%) \\
& \quad + KL & 62.5 (↓5.0\%) & 74.6 (↑3.5\%) & 67.6 (↑4.8\%) & 68.0 (↑2.7\%) \\
& \quad + JS & 65.8 (↑0.0\%) & 74.7 (↑3.6\%) & 67.6 (↑4.8\%) & 68.6 (↑3.6\%) \\
& \quad + RG & 58.3 (↓11.4\%) & 73.5 (↑1.9\%) & 67.4 (↑4.5\%) & 66.9 (↑1.1\%) \\
& \quad + ADV & 57.6 (↓12.5\%) & 74.2 (↑2.9\%) & 68.3 (↑5.9\%) & 67.5 (↑2.0\%) \\
& Ours & 62.6 (↓4.9\%) & 73.7 (↑2.2\%) & 68.4 (↑6.0\%) & \cellcolor{gray!20}\underline{68.4 (↑3.3\%)} \\
\bottomrule
\end{tabular}
}

\textit{Note.} While Qwen2.5-VL was originally instruction-tuned with proprietary in-house data~\citep{qwen25vl}, our reproduced version uses only publicly available LLaVA instruction-tuning data. Even under this constraint and without access to VQA-specific tuning samples, our models achieve comparable or even better performance across all VQA datasets—highlighting the robustness and effectiveness of our proposed perturbation-consistent fine-tuning strategy.

\vspace{-0.2cm}
\end{table*}

\clearpage

\begin{table*}[ht]
\centering
\small
\setlength{\tabcolsep}{3pt}
\renewcommand{\arraystretch}{1.1}
\caption{Performance (\%) of V a lin na models under modality interference across four datasets. We show accuracy under clean (origin) and various perturbations: Left: Mini-ImageNet and Caltech-101; Right: OpenBookQA and MMLU. UF = Unrelated Facts, MD = Misleading Descriptions, RP = Random Pixels, RI = Real Image, FB \& FW = Full Black/White Canvas. (The results are averaged on multiple runs with standard deviation $<$ 0.2) }
\resizebox{\textwidth}{!}{
\begin{tabular}{l|ccc|ccc|cccc|cccc}
\toprule
\multirow{2}{*}{Model} & \multicolumn{3}{c|}{Mini-ImageNet} & \multicolumn{3}{c|}{Caltech-101} & \multicolumn{4}{c|}{Open-Book QA} & \multicolumn{4}{c}{MMLU} \\
 & Orig & UF & MD & Orig & UF & MD & RP & RI & FB & FW & RP & RI & FB & FW \\
\midrule
InternVL2-2B~\citep{internvl2} & 91.9 & 88.6 & 25.5 & 94.6 & 91.3 & 33.2 & 46.3 & 36.2 & 45.2 & 45.3 & 39.1 & 36.9 & 39.3 & 39.2 \\
LLaVA-1.5-7B~\citep{llava15} & 95.3 & 93.4 & 43.5 & 97.0 & 95.9 & 57.4 & 62.4 & 56.4 & 62.5 & 63.4 & 46.3 & 45.2 & 45.9 & 45.8 \\
LLaVA-Next-7B~\citep{llavanext} & 93.4 & 90.0 & 28.5 & 97.0 & 93.4 & 31.6 & 54.6 & 52.9 & 55.9 & 55.5 & 45.9 & 45.0 & 45.9 & 45.8 \\
LLaVA-Next-34B~\citep{llavanext} & 98.0 & 96.3 & 90.2 & 99.1 & 97.3 & 93.6 & 87.7 & 85.7 & 88.4 & 88.0 & 71.5 & 70.9 & 71.5 & 71.5 \\
LLaVA-Next-72B~\citep{llavanext} & 97.7 & 97.8 & 83.0 & 98.9 & 98.8 & 91.2 & 88.2 & 86.2 & 89.1 & 88.9 & 73.6 & 72.9 & 73.9 & 73.9 \\
LLaVA-Next-110B~\citep{llavanext} & 98.4 & 98.3 & 93.3 & 98.4 & 98.3 & 93.1 & 89.5 & 89.2 & 89.9 & 89.7 & 73.5 & 73.0 & 73.9 & 73.9 \\
LLaVA-1.5-13B~\citep{llava15} & 95.6 & 94.1 & 73.0 & 97.9 & 97.1 & 77.4 & 65.9 & 63.8 & 68.0 & 69.1 & 51.8 & 50.8 & 52.7 & 52.7 \\
InstructBLIP-7B~\citep{instructblip} & 92.0 & 87.1 & 13.6 & 90.3 & 90.2 & 17.5 & 50.8 & 46.2 & 50.9 & 50.7 & 36.3 & 35.2 & 36.2 & 36.7 \\
InstructBLIP-13B~\citep{instructblip} & 93.0 & 81.6 & 50.9 & 94.1 & 82.7 & 51.0 & 44.7 & 39.4 & 45.6 & 46.1 & 40.1 & 37.4 & 40.5 & 42.9 \\
QwenVL2-2B~\citep{qwen2} & 98.7 & 98.6 & 75.8 & 99.1 & 99.1 & 66.8 & 59.6 & 54.7 & 63.4 & 63.8 & 49.5 & 44.6 & 49.7 & 50.0 \\
QwenVL2.5-3B~\citep{qwen25vl} & 98.9 & 98.5 & 94.9 & 98.8 & 99.0 & 94.4 & 79.9 & 74.6 & 80.0 & 79.7 & 63.5 & 61.1 & 64.0 & 63.6 \\
QwenVL3-4B~\citep{qwen3vl} & 99.6 & 99.1 & 96.4 & 99.3 & 99.2 & 98.0 & 84 & 82.3 & 86.7 & 86.4 & 71.8 & 70.9 & 72.4 & 72.4 \\
QwenVL2-7B~\citep{qwen2} & 99.0 & 99.1 & 96.3 & 99.6 & 99.5 & 97.6 & 82.5 & 80.9 & 83.7 & 83.2 & 66.9 & 65.3 & 67.7 & 67.8 \\
QwenVL2.5-7B~\citep{qwen25vl} & 99.3 & 99.3 & 96.5 & 99.3 & 99.4 & 96.8 & 86.2 & 80.9 & 86.3 & 86.4 & 69.5 & 67.6 & 69.1 & 69.1 \\
QwenVL3-8B~\citep{qwen3vl} & 99.3 & 98.8 & 97.2 & 99.8 & 99.7 & 98.1 & 87.0 & 85.7 & 87.5 & 88.1 & 75.6 & 75.0 & 76.2 & 76.3 \\
\bottomrule
\end{tabular}
}
\label{tab:modality_interference_vinilla_models}
\end{table*}

\begin{figure*}[t]
    \centering
    \includegraphics[width=\linewidth]{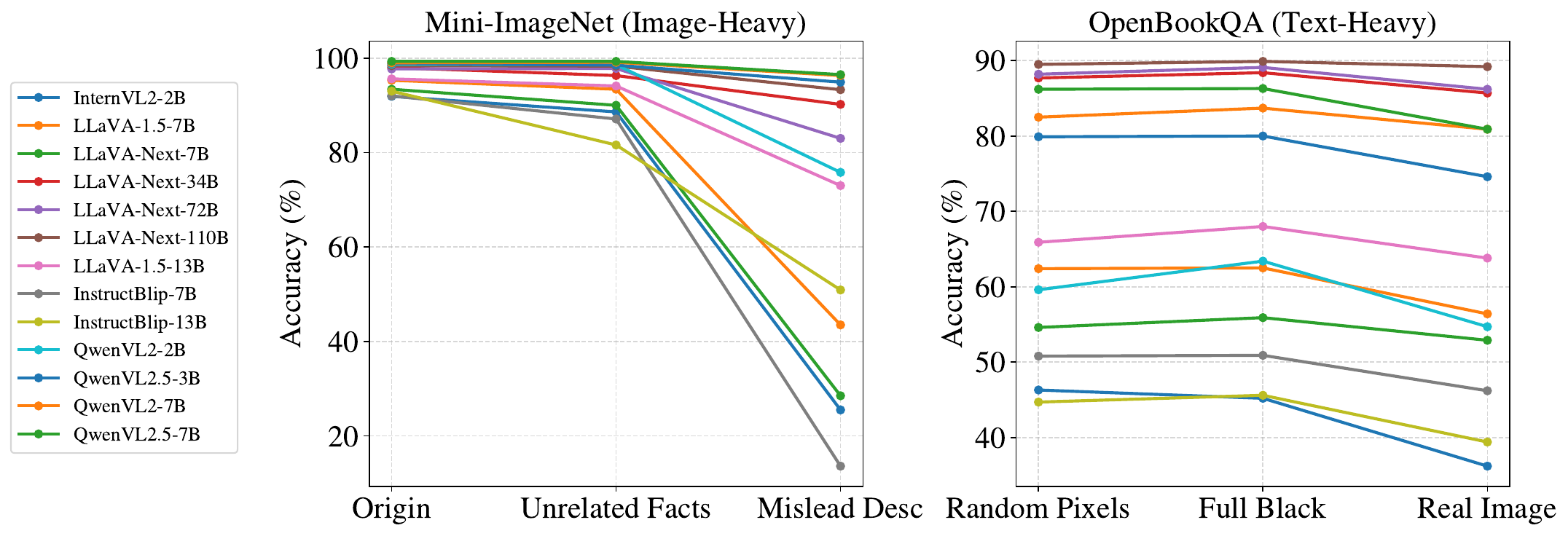}
    \caption{Performance degradation under irrelevant perturbations reveals modality interference in MLLMs. Left: Mini-ImageNet (image-heavy) with Original input, Unrelated Facts, and Misleading Descriptions. Right: OpenBookQA (text-heavy) with Random Pixels, Full Black Canvas, and Irrelevant Real Images. Misleading descriptions most severely affect image-heavy tasks, while irrelevant real images cause the largest drop in text-heavy reasoning.}
    \label{fig:pretrained_models_perturbation_results_line_chart}
\end{figure*}

\subsection{Perturbation-based Evaluation Experiment Results}
\label{apx:Perturbation-based Evaluation Experiment Results}
We conduct a controlled perturbation-based evaluation across various MLLMs, as shown in~\Cref{tab:modality_interference_vinilla_models}. Our results reveal that both vision and language tasks are vulnerable to cross-modal interference.
In vision classification tasks, misleading textual descriptions (e.g., text contradicting image content) lead to severe performance drops. For example, InternVL2-2B and InstructBLIP-7B on Mini-ImageNet drop from 91.9\% to 25.5\% and from 92.0\% to 13.6\%, respectively. Conversely, for language tasks such as OpenBookQA and MMLU, irrelevant visual inputs—particularly semantically unrelated real images—also degrade performance. LLaVA-1.5-7B drops from 46.3\% to 45.2\% on MMLU, while InstructBLIP-13B sees over 5 points of degradation.

A consistent trend is that larger models exhibit greater robustness. Models like LLaVA-1.5-13B and QwenVL2.5-7B maintain high accuracy across all perturbation types—e.g., QwenVL2.5-7B sustains over 96\% on Mini-ImageNet with misleading text and over 86\% on OpenBookQA with irrelevant images—indicating improved modality disentanglement and reduced sensitivity to spurious correlations. Nonetheless, performance still degrades relative to clean inputs, highlighting that interference effects remain non-negligible even in stronger models.

We further observe a clear scaling trend within the LLaVA-Next family. As model size increases from 7B to 34B, 72B, and 110B, performance under perturbations steadily improves, reflecting stronger representation power and enhanced robustness to spurious cues. For instance, LLaVA-Next-7B achieves 90.3\% on Mini-ImageNet (Orig) but drops to 28.5\% with misleading descriptions, whereas LLaVA-Next-110B maintains 98.4\% and 93.3\% under the same conditions. Similarly, on MMLU, accuracy under irrelevant real images increases from 45.9\% (7B) to 73.6\% (72B). \textbf{These results confirm that scaling up helps mitigate modality interference. However, such gains come at substantial computational and resource costs, and the improvements remain incremental relative to the clean–perturbed gap.} This underscores that scaling alone is insufficient, and more targeted interventions—such as our proposed framework—are necessary for robust cross-modal reasoning.

\subsection{Experiment Results on InstructBLIP-Vicuna-13b and QwenVL-2.5-7b}
To enable a more equitable comparison with existing multimodal models, we extend our method to two additional backbones: InstructBLIP-Vicuna-13B and QwenVL-2.5-7B. As shown in~\Cref{tab:vqa_summary_extra_models}, our approach consistently improves performance across multiple VQA benchmarks, even under different instruction-tuning conditions, demonstrating its robustness and general applicability.

\subsection{Discussion on KL\&JS use for Consistency Regularization}
Although both KL and JS divergence serve as effective objectives for consistency regularization, we find that JS consistently achieves slightly better results across most settings. 
Specifically, in both unimodal tasks (e.g., Mini-ImageNet, MMLU) and multimodal reasoning benchmarks (e.g., ScienceQA, SeedBench), JS-regularized models consistently outperform their KL counterparts by a small but observable margin. This trend holds across different model backbones and training configurations, including our final unified method (see “Ours” rows in Table~\ref{tab:unimodal_summary} and~\ref{tab:vqa_summary_extra_models}).
This suggests a marginal advantage of JS regularization in enhancing model robustness.

\subsection{Evaluating Chain-of-Thought Prompting for Modality Interference Mitigation}
\label{sec:cot}

To further investigate the potential of prompt-based methods in mitigating modality interference, we conduct additional experiments using Chain-of-Thought \cite{cot} style prompting. This approach aims to encourage structured reasoning by guiding the model through an explicit reasoning process before producing its final answer.

Specifically, we prepend the following CoT prompt to each input question:

\begin{quote}
\small
\texttt{Let’s think step by step:} \\
\texttt{\ \ \ 1.\quad What information does the image provide?} \\
\texttt{\ \ \ 2.\quad What is the question asking?} \\
\texttt{\ \ \ 3.\quad Are there any misleading parts?} \\
\texttt{\ \ \ 4.\quad Now give your final answer. Only write the final answer on a separate line like: ``Answer: B''}
\end{quote}

Results are presented in Table~\ref{tab:cot-results}. The results suggests that structured reasoning alone cannot resolve the interference problem, as the issue stems from misaligned cross-modal representations rather than shallow reasoning steps.

\begin{table*}[ht]
\centering
\scriptsize
\setlength{\tabcolsep}{6pt}
\caption{Accuracy (\%) under different interference settings across tasks and models. Each task includes original inputs and multiple types of perturbations.}
\label{tab:cot-results}
\begin{tabular}{l|l|ccc|ccc|cccc|cccc}
\toprule
\multirow{2}{*}{\textbf{Model}} & \multirow{2}{*}{\textbf{Method}} &
\multicolumn{3}{c|}{\textbf{Mini-ImageNet}} &
\multicolumn{3}{c|}{\textbf{Caltech-101}} &
\multicolumn{4}{c|}{\textbf{OpenBookQA}} &
\multicolumn{4}{c}{\textbf{MMLU}} \\
& & Orig & UF & MD & Orig & UF & MD & RP & RI & FB & FW & RP & RI & FB & FW \\
\midrule
vanilla & llava-1.5-7b & 95.3 & 93.4 & 43.5 & 97.0 & 95.9 & 57.4 & 62.4 & 56.4 & 62.5 & 63.4 & 46.3 & 45.2 & 45.9 & 45.8 \\
CoT     & llava-1.5-7b & 81.7 & 70.9 & 28.9 & 80.9 & 71.8 & 36.0 & 38.8 & 38.9 & 41.0 & 41.0 & 39.6 & 38.7 & 40.2 & 40.4 \\
Ours    & llava-1.5-7b & 98.6 & 98.4 & 98.7 & 99.3 & 98.9 & 99.3 & 81.8 & 81.0 & 81.7 & 81.7 & 51.5 & 50.9 & 51.5 & 51.4 \\
\midrule
vanilla & llava-1.5-13b & 95.6 & 94.1 & 73.0 & 97.9 & 97.1 & 77.4 & 65.9 & 63.8 & 68.0 & 69.1 & 51.8 & 50.8 & 52.7 & 52.7 \\
CoT     & llava-1.5-13b & 92.9 & 85.9 & 62.8 & 96.6 & 92.1 & 67.8 & 55.6 & 53.0 & 56.5 & 57.3 & 47.3 & 45.5 & 47.3 & 47.8 \\
Ours    & llava-1.5-13b & 98.7 & 97.9 & 98.0 & 98.7 & 98.7 & 98.8 & 83.2 & 81.2 & 83.8 & 83.0 & 52.2 & 51.6 & 52.3 & 53.4 \\
\midrule
vanilla & qwen2.5-vl-3b & 98.9 & 98.5 & 94.9 & 98.8 & 99.0 & 94.4 & 79.9 & 74.6 & 80.0 & 79.7 & 63.5 & 61.1 & 64.0 & 63.6 \\
CoT     & qwen2.5-vl-3b & 92.3 & 96.8 & 88.2 & 94.7 & 96.1 & 86.0 & 61.0 & 52.8 & 61.3 & 61.1 & 51.5 & 48.7 & 51.6 & 51.4 \\
Ours    & qwen2.5-vl-3b & 99.3 & 99.2 & 99.2 & 99.7 & 99.7 & 99.5 & 86.7 & 86.4 & 86.6 & 86.6 & 64.8 & 64.5 & 65.0 & 65.1 \\
\midrule
vanilla & qwen2.5-vl-7b & 99.3 & 99.3 & 96.3 & 99.1 & 98.9 & 97.2 & 85.9 & 77.5 & 85.8 & 86.0 & 69.3 & 63.7 & 68.9 & 68.9 \\
CoT     & qwen2.5-vl-7b & 99.1 & 98.9 & 91.4 & 98.5 & 98.5 & 92.5 & 77.8 & 68.8 & 77.3 & 77.6 & 61.5 & 57.4 & 60.8 & 60.9 \\
Ours    & qwen2.5-vl-7b & 99.6 & 99.5 & 99.5 & 99.6 & 99.6 & 99.7 & 90.9 & 89.7 & 90.8 & 91.0 & 69.8 & 68.2 & 69.8 & 70.0 \\
\bottomrule
\end{tabular}
\end{table*}

\subsection{Evaluation on Free-Form VQA}
\label{sec:freeformVQA}
To further assess the generalizability of our method beyond multiple-choice VQA tasks, we evaluate on TextVQA~\citep{textvqa}, a free-form generative visual question answering dataset that requires reasoning over both textual and visual content in natural images. We follow the benchmark adopted by LLaVA \cite{llava}, which evaluates a model’s ability to both recognize textual characters within images and effectively handle noisy outputs generated by OCR systems.  

Following standard evaluation protocols and existing MLLM baselines (e.g., LLaVA, Qwen2.5-VL, InstructBLIP), we report model performance averaged over multiple runs (standard deviation $<$ 0.4). Results are presented in Table~\ref{tab:textvqa}.

Our method achieves consistent improvements across most model families, indicating its effectiveness not only in MCQA scenarios but also in open-ended multimodal reasoning settings. 

\begin{table}[ht]
\centering

\caption{Accuracy (\%) on the TextVQA dataset across different model families.}
\label{tab:textvqa}
\resizebox{0.4\textwidth}{!}{
\begin{tabular}{llc}
\toprule
\textbf{Setting} & \textbf{Model} & \textbf{TextVQA} \\
\midrule
Vanilla & InstructBLIP-Vicuna-7B & 19.7 \\
Ours    & InstructBLIP-Vicuna-7B & \textbf{32.4} \\
\midrule
Vanilla & LLaVA-1.5-7B           & 49.8 \\
Ours    & LLaVA-1.5-7B           & \textbf{51.2} \\
\midrule
Vanilla & Qwen2.5-VL-7B          & 81.4 \\
Ours    & Qwen2.5-VL-7B          & \textbf{84.8} \\
\bottomrule
\end{tabular}
}
\end{table}

\subsection{Improving Generalizability through Adversarial Training}

To assess the generalization benefits of adversarial training, we evaluate model robustness under two types of out-of-distribution (OOD) perturbations at test time:

\begin{itemize}
    \item \textbf{Document OCR noise:} Real-world noisy OCR snippets are sampled from the FUNSD dataset~\citep{funsd} and inserted as irrelevant textual distractors into image-heavy visual classification tasks (Mini-ImageNet, Caltech-101).
    \item \textbf{Unrelated screenshots:} Unrelated UI screenshots are drawn from the RICO dataset~\citep{rico} and added as visual distractors to text-heavy reasoning tasks (OpenBookQA, MMLU).
\end{itemize}

Beyond these seen datasets, we further evaluate generalization on two \emph{completely unseen datasets} at test time: \textbf{Food-101} and \textbf{ARC Challenge}. Food-101 serves as an unseen image-heavy classification benchmark, where models are exposed to diverse textual perturbations, including unrelated descriptions, misleading class cues, and OCR-style noise. ARC Challenge represents an unseen text-heavy reasoning task, where various visual distractors—such as random pixels, unrelated real images, full-black/white images, and UI screenshots—are introduced to probe robustness against spurious visual signals.

Each experiment is repeated multiple times, and we report average accuracy across runs (standard deviation $< 0.1$). Results on seen datasets are presented in Table~\Cref{tab:ood-results-new-perturbations}, while results on unseen datasets are summarized in ~\Cref{tab:ood-results-new-datasets}. Across all model families and task types, adversarial training consistently improves robustness to both unseen perturbation types and unseen datasets. In contrast, heuristic data augmentation alone often fails to generalize beyond the perturbation patterns observed during training and can even degrade performance under novel noise.

Overall, these results demonstrate that adversarial training substantially enhances out-of-domain generalization with only modest additional training overhead. This improved robustness to distribution shifts is crucial for reliable deployment of multimodal models in real-world, noisy environments.

\begin{table*}[ht]
\centering
\caption{Accuracy (\%) on original and perturbed inputs. OCR snippets are inserted into image classification tasks, and RICO screenshots are added to VQA tasks.}
\label{tab:ood-results-new-perturbations}
\scriptsize
\begin{tabular}{llcccccccc}
\toprule
\textbf{Setting} & \textbf{Model} & \multicolumn{2}{c}{Mini-ImageNet} & \multicolumn{2}{c}{Caltech-101} & \multicolumn{2}{c}{OpenBookQA} & \multicolumn{2}{c}{MMLU} \\
 &  & Origin & OCR & Origin & OCR & RandPixels & Screenshot & RandPixels & Screenshot \\
\midrule
vanilla   & instructblip-vicuna-7b & 92.0 & 81.4 & 90.3 & 83.6 & 50.9 & 40.0 & 35.3 & 34.9 \\
SFT       & instructblip-vicuna-7b & 98.5 & 95.7 & 99.0 & 98.4 & 75.0 & 74.4 & 50.0 & 49.2 \\
SFT + ADV & instructblip-vicuna-7b & 98.7 & 97.5 & 99.5 & 98.8 & 76.8 & 76.2 & 49.3 & 49.3 \\
Ours      & instructblip-vicuna-7b & 98.4 & 97.3 & 99.2 & 99.0 & 79.0 & 77.8 & 50.2 & 49.2 \\
\midrule
vanilla   & llava-1.5-7b           & 95.3 & 88.9 & 97.0 & 92.8 & 62.4 & 50.2 & 46.3 & 44.8 \\
SFT       & llava-1.5-7b           & 98.2 & 98.0 & 98.5 & 98.0 & 78.6 & 78.0 & 51.1 & 50.6 \\
SFT + ADV & llava-1.5-7b           & 98.7 & 98.2 & 98.7 & 98.4 & 81.7 & 81.3 & 50.6 & 51.3 \\
Ours      & llava-1.5-7b           & 98.6 & 98.2 & 99.3 & 99.0 & 81.8 & 81.2 & 51.5 & 51.3 \\
\midrule
vanilla   & qwen2.5-vl-7b          & 99.3 & 99.2 & 99.1 & 99.2 & 85.9 & 69.3 & 69.3 & 57.0 \\
SFT       & qwen2.5-vl-7b          & 99.6 & 99.5 & 99.7 & 99.3 & 90.2 & 85.8 & 70.4 & 63.7 \\
SFT + ADV & qwen2.5-vl-7b          & 99.4 & 99.5 & 99.6 & 99.6 & 91.7 & 91.5 & 70.4 & 69.7 \\
Ours      & qwen2.5-vl-7b          & 99.6 & 99.5 & 99.6 & 99.5 & 90.9 & 90.7 & 69.8 & 69.7 \\
\bottomrule
\end{tabular}
\end{table*}

\begin{table*}[ht]
\centering
\caption{Accuracy (\%) on Food-101 (image-heavy) and ARC Challenge (text-heavy) under diverse test-time perturbations. For Food-101, we add different types of textual distractors (unrelated, misleading, and OCR-style noise). For ARC Challenge, we use various visual distractors (random pixels, unrelated real images, full-black/white images, and UI screenshots).}
\label{tab:ood-results-new-datasets}
\scriptsize
\begin{tabular}{llcccc|ccccc}
\toprule
\textbf{Setting} & \textbf{Model} &
\multicolumn{4}{c|}{\textbf{Food-101}} &
\multicolumn{5}{c}{\textbf{ARC Challenge}} \\
& &
Origin & Unrelated Text & Mislead Text & OCR Text &
RandPixels & Real Image & Full Black & Full White & Screenshot \\
\midrule
vanilla        & qwen2.5-vl-7b          & 97.1 & 96.8 & 93.3 & 96.7 & 85.6 & 83.7 & 85.9 & 85.9 & 79.6 \\
+$D^{\text{AUG}}$ & qwen2.5-vl-7b       & 97.1 & 96.6 & 94.1 & 96.5 & 85.0 & 82.9 & 84.9 & 84.9 & 80.2 \\
Ours           & qwen2.5-vl-7b          & 97.5 & 97.5 & 97.6 & 97.4 & 84.1 & 82.7 & 84.2 & 84.0 & 82.6 \\
\midrule
vanilla        & llava-1.5-7b           & 90.2 & 86.3 & 31.0 & 83.6 & 53.4 & 51.6 & 54.5 & 54.0 & 51.8 \\
+$D^{\text{AUG}}$ & llava-1.5-7b        & 91.3 & 86.5 & 40.2 & 83.5 & 55.7 & 52.0 & 56.0 & 56.0 & 53.8 \\
Ours           & llava-1.5-7b           & 92.1 & 89.7 & 69.0 & 89.5 & 62.8 & 62.0 & 62.6 & 62.8 & 62.6 \\
\midrule
vanilla        & instructblip-vicuna-7b & 67.9 & 20.1 & 18.7 & 48.0 & 33.7 & 35.8 & 33.7 & 31.2 & 36.5 \\
+$D^{\text{AUG}}$ & instructblip-vicuna-7b & 55.5 & 41.4 & 38.4 & 56.5 & 27.3 & 28.2 & 27.6 & 27.2 & 30.7 \\
Ours           & instructblip-vicuna-7b & 95.6 & 89.8 & 93.7 & 88.1 & 57.6 & 56.4 & 58.8 & 59.6 & 57.0 \\
\bottomrule
\end{tabular}
\end{table*}

\subsection{Ablation without Modality Mask}
\label{apx:ablation_no_mask}

To further examine the role of modality-aware masking, we implement a \emph{No Mask} variant, where adversarial perturbations are applied uniformly to all modalities across all datasets, including VQA tasks.

~\Cref{tab:ablation_no_mask} reports the results on representative image-heavy and text-heavy benchmarks.
We observe three key findings.
First, training without a modality mask still partially mitigates modality interference.
This is because the combination of consistency regularization and heuristic perturbations already encourages the model to learn certain modality-invariant features, and adversarial training itself improves generic robustness (e.g., on OpenBookQA with screenshot distractors, accuracy improves from 50.2\% to 62.7\%).

Second, the modality mask is critical for out-of-distribution generalization.
Without masking, performance collapses on unseen perturbations, particularly under strong visual distractions.
For example, on OpenBookQA with screenshot perturbations, the No Mask variant underperforms the masked version by a large margin (62.7\% vs.\ 81.8\%).

Third, modality-aware masking is essential to preserve multimodal semantics during VQA training.
When adversarial noise is injected into all token embeddings indiscriminately, the model fails to maintain coherent cross-modal representations, resulting in a notable degradation in overall VQA performance (61.5\% vs.\ 66.8\%).

Together, these results demonstrate that while adversarial training alone improves robustness, modality-aware masking plays a crucial role in preventing semantic destruction and enabling reliable generalization under unseen distribution shifts. 

\begin{table*}[ht]
\centering
\caption{Ablation study without modality-aware masking.
The \emph{No Mask} variant applies adversarial perturbations to all modalities across all datasets.
Results are reported on image-heavy (Caltech-101) and text-heavy (OpenBookQA) benchmarks, together with overall VQA accuracy.}
\label{tab:ablation_no_mask}
\scriptsize
\begin{tabular}{lcccccc|c}
\toprule
\textbf{Model} &
\multicolumn{3}{c}{\textbf{Caltech-101}} &
\multicolumn{3}{c}{\textbf{OpenBookQA}} &
\textbf{VQA} \\
&
Orig & Mislead Text & OCR Text &
RandPixels & Real Image & Screenshot &
Overall \\
\midrule
LLaVA-1.5-7B (Vanilla)
& 97.0 & 57.4 & 92.8
& 62.4 & 56.4 & 50.2
& 63.8 \\
LLaVA-1.5-7B (Ours, No Mask)
& 98.1 & 96.3 & 93.5
& 79.5 & 78.1 & 62.7
& 61.5 \\
LLaVA-1.5-7B (Ours)
& \textbf{99.3} & \textbf{98.9} & \textbf{99.0}
& \textbf{81.8} & \textbf{81.0} & \textbf{81.8}
& \textbf{66.8} \\
\bottomrule
\end{tabular}
\end{table*}

\subsection{Evaluation on Reasoning-Enhanced (``Thinking'') Models}
\label{apx:thinking_models}

Reasoning-enhanced ``thinking'' models may potentially mitigate modality interference by explicitly reasoning over multimodal inputs.
To partially examine this hypothesis, we first evaluate chain-of-thought (CoT) prompting in the main paper (Table~1).
While CoT prompting reliably improves reasoning accuracy on clean inputs, it does not reduce modality interference under modality-mismatched perturbations, and the same failure modes persist.

To further study this question, we additionally evaluate \textbf{R1-OneVision-7B}~\citep{r1onevision}, a multimodal large language model explicitly designed with enhanced reasoning capabilities with a backbone Qwen2.5-VL-7B model.
We compare its performance against a strong non-reasoning baseline, \textbf{Qwen2.5-VL-7B}, under representative image-heavy and text-heavy OOD settings.

~\Cref{tab:thinking_model_results} reports the results.
Despite its stronger reasoning ability, R1-OneVision-7B still exhibits substantial performance degradation under misleading textual perturbations and visual distractors.
In contrast, Qwen2.5-VL-7B shows comparable or stronger robustness under the same conditions.

These results suggest that improved reasoning alone is insufficient to resolve modality interference.
Without explicit mechanisms to control modality reliance, such as modality-aware perturbation and regularization, even reasoning-enhanced models remain vulnerable to task-irrelevant signals.

We investigated the generated reasoning traces: In some cases the model shows better awareness of irrelevant modalities, but in many cases the model still treats unrelated modality as meaningful evidence and produces long hallucinated interpretations that directly influence the final answer. This confirms that our methods remain necessary for interference mitigation.

\begin{table*}[ht]
\centering
\caption{Evaluation of reasoning-enhanced ``thinking'' models under modality-mismatched perturbations.
Results are reported on image-heavy (Caltech-101) and text-heavy (OpenBookQA) tasks.}
\label{tab:thinking_model_results}
\scriptsize
\begin{tabular}{lcccc}
\toprule
\textbf{Model} &
\textbf{Caltech-101 (Orig)} &
\textbf{Caltech-101 (Mislead Text)} &
\textbf{OpenBookQA (RandPixels)} &
\textbf{OpenBookQA (Real Image)} \\
\midrule
R1-OneVision-7B
& 96.5 & 84.7 & 81.2 & 77.6 \\
Qwen2.5-VL-7B
& \textbf{99.1} & \textbf{97.2} & \textbf{85.9} & \textbf{77.5} \\
\bottomrule
\end{tabular}
\end{table*}

\clearpage

\section{Hyper-Parameter Setting and Training details}
\label{Apx:Parameter Setting}
Please see ~\Cref{tab:hyperparams} for more details. 

\begin{table}[t]
\centering
\begin{minipage}[t]{0.48\textwidth}
\centering
\caption{Hyperparameter Settings Example}
\label{tab:hyperparams}
\small
\begin{tabular}{l|l}
\toprule
\textbf{Category} & \textbf{Setting} \\
\midrule
\multicolumn{2}{c}{\textit{Model and Training Strategy}} \\
\midrule
Base Model & LLaVA-1.5-7B \\
Finetune Type & Full \\
Adversarial Type & PGD-alike \\
Step size $\alpha$ & 0.1 \\
Epsilon-hall $\epsilon$ & 0.001 \\
Avdersarial Training Steps $T$ & 2 \\
Consistency Regularization Type & JS \\
Loss Weight $\lambda_{\text{consistency}}$& 0.01 \\
Temperature $\tau$ & 1 \\
\midrule
\multicolumn{2}{c}{\textit{Optimization}} \\
\midrule
Epochs & 1 \\
Batch Size per GPU $|\mathcal{B}|$ & 8 \\
Img/Text Ratio & 0.25/0.25 \\
Learning rate & $2 \times 10^{-5}$ \\
\bottomrule
\end{tabular}
\end{minipage}
\hfill
\begin{minipage}[t]{0.48\textwidth}
\centering
\caption{Dataset Statistics}
\label{tab:dataset_sizes}
\begin{tabular}{lrr}
\toprule
\textbf{Dataset} & \textbf{Train} & \textbf{Test} \\
\midrule
Mini-ImageNet & 4935 & 2000 \\
Caltech-101 & 8124 & 1020 \\
OpenBookQA & 4957 & 1000 \\
MMLU & 7M & 5469 \\
LLaVA-Instruct & 624610 & - \\
TextCaps & 109765 & - \\
MMBench-EN & - & 4377 \\
ScienceQA-IMG & - & 4114 \\
SeedBench-IMG & - & 14243 \\
\bottomrule
\end{tabular}
\end{minipage}
\end{table}

\subsection{Parameter Analysis with adversarial training iterations}

\begin{figure*}[t]
  \centering
  \begin{minipage}[t]{0.8\textwidth}
    \centering
    \includegraphics[width=\linewidth]{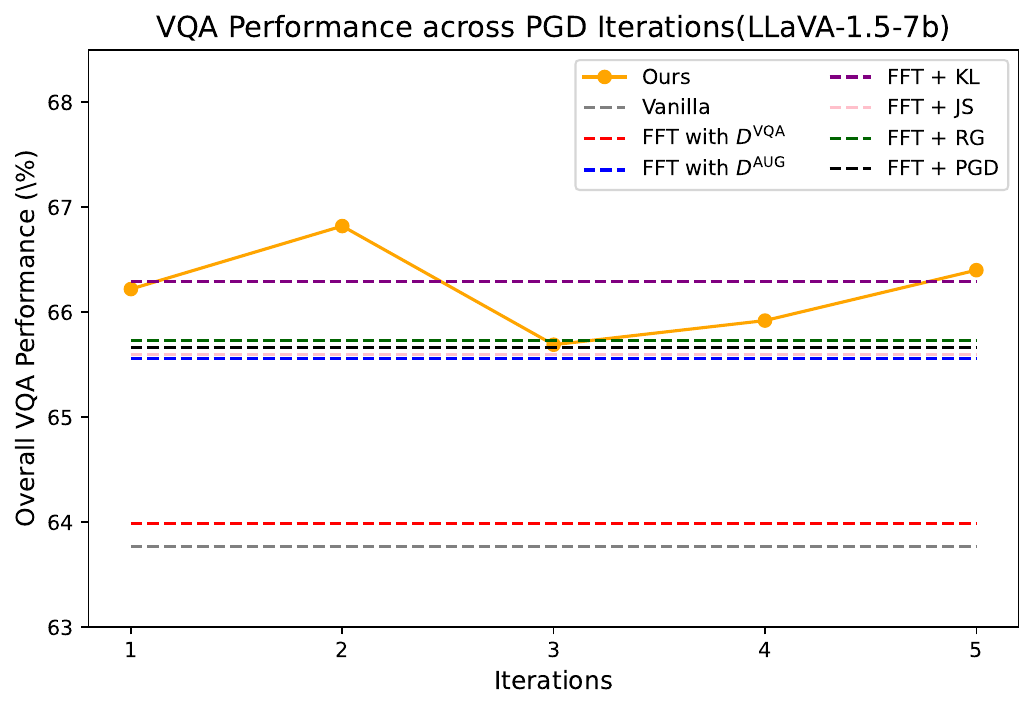}
  \end{minipage}
  \hfill
  \begin{minipage}[t]{0.8\textwidth}
    \centering
    \includegraphics[width=\linewidth]{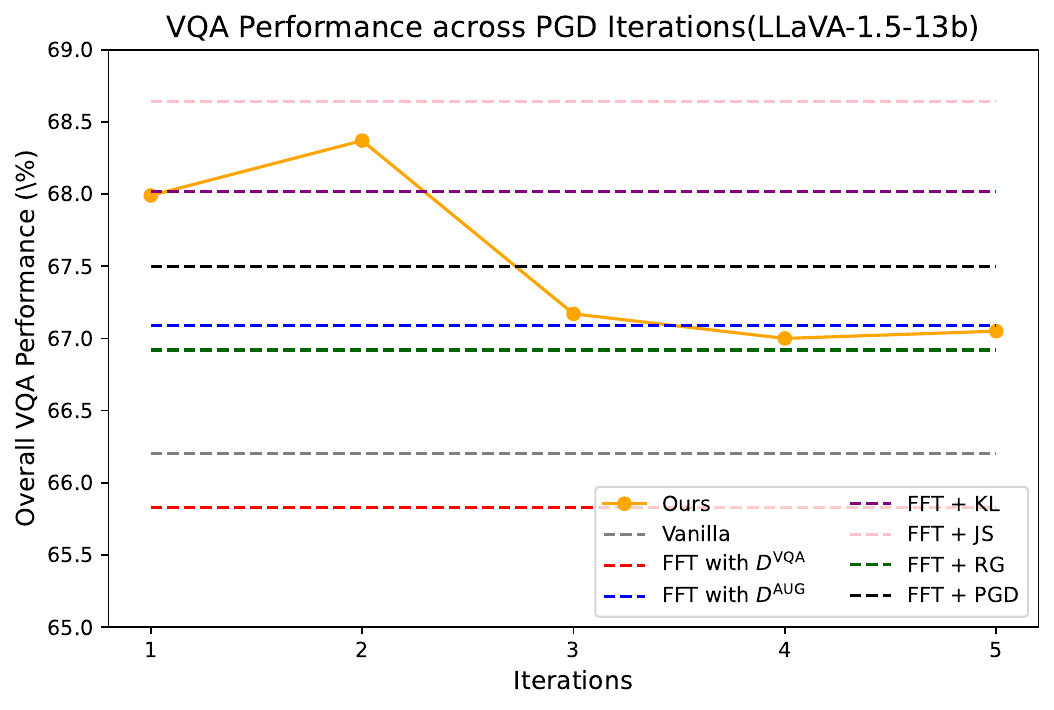}
  \end{minipage}
  \caption{Comparison of VQA performance across adversarial training iterations for different model sizes.}
  \label{fig:llava_pgd_compare}
\end{figure*}

To investigate the effect of adversarial strength on model performance, we vary the number of adversarial training iterations from 1 to 5 and evaluate the resulting VQA accuracy. As shown in ~\Cref{fig:llava_pgd_compare}, both LLaVA-1.5-7B and LLaVA-1.5-13B models benefit from adversarial consistency training, with performance peaking at 2-step adversarial training (66.82\% and 68.37\%, respectively). Notably, excessive iterations (e.g., 4 or 5 steps) may lead to slight degradation, especially in larger models, likely due to over-perturbation and optimization difficulty. 

These findings suggest that a moderate adversarial training setting (2 steps with $\epsilon$=1e-3 and $\alpha$=0.1 in LLaVA-1.5-7b, $\epsilon$=1e-4 and $\alpha$=0.1 in LLaVA-1.5-13b) offers an optimal balance between robustness and training stability, and that model size influences sensitivity to adversarial signal strength.

\clearpage
\section{Experiments compute resources}
\label{Apx:Implementation Details}
All experiments were conducted on 8$\times$A100 GPUs using DeepSpeed ZeRO-3~\citep{deepspeed2024} with CPU offloading.

To quantify the computational overhead introduced by our adversarial training, we provide both theoretical FLOPs analysis and empirical wall-clock training time across model scales.

In the standard supervised fine-tuning (SFT) setting, the FLOPs per batch can be approximated as:
\begin{equation}
\text{FLOPs}_{\text{SFT}} \approx B_s \cdot (f_{\text{LLM}} + b_{\text{LLM}}),
\end{equation}
where $B_s$ is the batch size, $f_{\text{LLM}}$ denotes the FLOPs of a forward pass through the LLM, and $b_{\text{LLM}}$ the backward pass.
In our adversarial training, each sample undergoes $N$ adversarial training steps, each requiring an additional forward pass through the \emph{frozen} LLM. Since gradients are computed only with respect to input embeddings via \texttt{torch.autograd.grad}, the overhead is minimal and thus ignored. After perturbation, clean and adversarial inputs are concatenated, resulting in a forward cost of $2B_s \cdot f_{\text{LLM}}$, followed by one backward pass. The total cost becomes:
\begin{equation}
\text{FLOPs}_{\text{ADV+SFT}} = B_s \cdot (N f_{\text{LLM}} + 2 f_{\text{LLM}} + b_{\text{LLM}}).
\end{equation}

The relative overhead compared to vanilla SFT is:
\begin{equation}
\frac{N f + 2 f + b}{f + b}.
\end{equation}

Assuming $b_{\text{LLM}} \approx 2f_{\text{LLM}}$, this simplifies to:
\begin{equation}
\frac{N + 4}{3}.
\end{equation}

For our default $N = 1$, the theoretical FLOPs increase to approximately $1.66\times$ that of SFT.

We further report the actual training time across model scales, as shown in Table~\ref{tab:training-time}.

\begin{table}[h]
\centering
\caption{Training time (in hours) across model scales. $\Delta$ denotes additional overhead per adversarial training iteration step.}
\label{tab:training-time}
\resizebox{0.45\textwidth}{!}{
\begin{tabular}{lcccc}
\toprule
\textbf{Model} & \textbf{SFT} & \textbf{SFT + KL} & \textbf{SFT + ADV} & \textbf{Ours (ADV + KL)} \\
\midrule
Qwen2.5-VL-3B  & 1.5h         & 1.5--1.75h        & 3--3.5h ($\Delta$ = 0.5h) & 3.5--4h ($\Delta$ = 0.5h) \\
LLaVA-1.5-7B   & 4h           & 4--4.5h           & 6--6.5h ($\Delta$ = 0.5h) & 6--6.5h ($\Delta$ = 0.5h) \\
LLaVA-1.5-13B  & 10h          & 10--11h           & 13--14h ($\Delta$ = 1h)   & 13--14h ($\Delta$ = 1h)   \\
\bottomrule
\end{tabular}
}
\end{table}

Although the theoretical FLOPs suggest a $\sim$66\% increase in cost when $N = 1$, the actual wall-clock time increase is much smaller. This is because our designed adversarial training leverages forward-only passes over frozen LLMs, avoiding costly backward and optimizer updates. As a result, the added runtime remains modest even on large models (e.g., only +2.5h for LLaVA-13B). Moreover, KL consistency training introduces negligible overhead compared to SFT.

\section{Limitations}
\label{Apx:Limitations}
Our analysis of modality interference is conducted from a coarse-grained perspective, primarily categorizing tasks into image-heavy and text-heavy types. A more fine-grained investigation—such as dynamic attention tracking—could provide deeper insights into how MLLMs rely on or ignore specific modalities during reasoning.

Moreover, while our perturbation strategies (e.g., unrelated facts, misleading descriptions, irrelevant images) effectively reveal failure modes of current MLLMs, they remain heuristic and task-specific. Designing perturbations is, by nature, an open-ended process—one can always propose new forms of misleading inputs. Thus, an ultimate goal is to develop perturbation-agnostic methods that improve robustness without requiring exhaustive enumeration of possible attacks.

While our use of adversarial training represents a strong and generalizable perturbation strategy, it still operates within a defined input space (e.g., embedding-level noise bounded by $L_\infty$ norms). Hence, adversarial perturbation should be viewed as a practical but partial solution rather than a comprehensive defense. Developing mechanisms that generalize across both semantic and modality perturbations remains an open and challenging direction.

\section{The Use of Large Language Models}
\label{Apx:LLM_use}
We used large language models only to edit the manuscript for clarity, grammar, and academic style. No part of the research design, data analysis, or scientific content relied on language models, and the authors retain full responsibility for the paper’s ideas and conclusions.

\end{document}